\documentclass[journal=jacsat,manuscript=article]{achemso}

\usepackage[version=3]{mhchem} % Formula subscripts using \ce{}
\usepackage{subfigure}
\usepackage{amsmath}
\usepackage{graphicx}
\usepackage{comment}
\usepackage{color}
\usepackage{lineno}
\usepackage{siunitx}
\usepackage{longtable}
\newcommand\Tg{$T_\text{g}\;$}
\mciteErrorOnUnknownfalse

\author{Pranav Shetty}
\affiliation[1]{School of Computational Science \& Engineering}
\author{Arunkumar Chitteth Rajan}
\author{Christopher Kuenneth}
\affiliation[2]{School of Materials Science and Engineering, Georgia Institute of Technology, 771 Ferst Drive NW, Atlanta, Georgia 30332, USA}
\author{Sonakshi Gupta}
\affiliation[3]{Department of Metallurgy Engineering and Materials Science, Indian Institute of Technology, Indore}
\author{Lakshmi Prerana Panchumarti}
\affiliation[2]{School of Materials Science and Engineering, Georgia Institute of Technology, 771 Ferst Drive NW, Atlanta, Georgia 30332, USA}
\author{Lauren Holm}
\affiliation[2]{School of Materials Science and Engineering, Georgia Institute of Technology, 771 Ferst Drive NW, Atlanta, Georgia 30332, USA}
\author{Chao Zhang}
\affiliation[1]{School of Computational Science \& Engineering}
\author{Rampi Ramprasad}
\affiliation[2]{School of Materials Science and Engineering, Georgia Institute of Technology, 771 Ferst Drive NW, Atlanta, Georgia 30332, USA}
\email{rampi.ramprasad@mse.gatech.edu}

\title[A general-purpose material property data extraction pipeline from large polymer corpora using Natural Language Processing]
{A general-purpose material property data extraction pipeline from large polymer corpora using Natural Language Processing}

\begin{document}


\begin{abstract}

The ever-increasing number of materials science articles makes it hard to infer chemistry-structure-property relations from published literature. We used natural language processing (NLP) methods to automatically extract material property data from the abstracts of polymer literature. As a component of our pipeline, we trained MaterialsBERT, a language model, using 2.4 million materials science abstracts, which outperforms other baseline models in three out of five named entity recognition datasets when used as the encoder for text. Using this pipeline, we obtained $\sim$300,000 material property records from $\sim$130,000 abstracts in 60 hours. The extracted data was analyzed for a diverse range of applications such as fuel cells, supercapacitors, and polymer solar cells to recover non-trivial insights. The data extracted through our pipeline is made available through a web platform at \url{https://polymerscholar.org} which can be used to locate material property data recorded in abstracts conveniently. This work demonstrates the feasibility of an automatic pipeline that starts from published literature and ends with a complete set of extracted material property information.

\end{abstract}

\section{Introduction}

The number of materials science papers published annually grows at the rate of \SI{6}{\%} compounded annually. Quantitative and qualitative materials property information is locked away in these publications written in natural language that is not directly machine-readable. The explosive growth in published literature makes it harder to see quantitative trends by just directly analyzing large amounts of literature. Searching the literature for material systems that have desirable properties also becomes more challenging. Moreover, material information published in a non-machine-readable form contributes to data scarcity in the field of materials informatics where the training of property predictors requires painstaking manual curation of the data of interest from literature. Here, we propose adapting techniques for information extraction from the natural language processing (NLP) literature to address these issues.

Information extraction from the written text is well studied within NLP and
involves several key components such as named entity recognition (NER),
i.e., identifying categories to which words in the text belong; relation
extraction, i.e., classifying relationships between extracted entities;
co-referencing, i.e., identifying clusters of named entities in the text
referring to the same object such as a polymer and its abbreviation, and named entity normalization, i.e. identifying all the variations in the name 
for an entity across a large number of documents. The idea
of ``self-supervised learning'' through transformer-based models such as BERT \cite{devlin2018bert, vaswani2017attention}, pre-trained on massive corpora of unlabeled text to learn contextual embeddings, is the dominant paradigm of information extraction today. A common architecture for NER and relation extraction is to feed a labeled input to BERT and use the output vector embedding for each word along with the corresponding labels (which could be entity labels or relation labels) as inputs to a downstream machine learning
model (typically a neural network) that learns to predict those labels. The tasks mentioned above
are label intensive. Extending these methods to new domains requires labeling new data sets with ontologies that are tailored to the specific domain of interest.

ChemDataExtractor \cite{chemdataextractor}, ChemSpot \cite{chemspot}, and ChemicalTagger \cite{chemicaltagger} are tools that perform NER to tag material entities.
For example, ChemDataExtractor has been used to create a database of Neel temperatures and Curie temperatures that were automatically mined from literature \cite{court2018auto}.
It has also been used to generate a literature extracted database of magnetocaloric materials and train property prediction models for key figures of merit \cite{court2021inverse}.
In the space of polymers, the authors of Ref. \citenum{tchoua2016blending} used a semi-automated approach that crawled papers automatically and used students to extract the Flory-Huggins parameter (a measure of the affinity between two materials, eg., a polymer and a solvent).
Word embedding approaches were used in Ref. \citenum{tchoua2019creating} to generate entity-rich documents for human experts to annotate which were then used to train a polymer named entity tagger. Most previous NLP-based efforts in materials science have focused on inorganic materials but limited work has been done to address information extraction challenges in polymers. Polymers in practice have several non-trivial variations in name for the same material entity which requires polymer names to be normalized. Moreover, unlike inorganic entities, polymer names cannot typically be converted to SMILES strings that are usable for downstream machine learning but the SMILES strings must instead be inferred from figures in the paper that contain the corresponding structure.

Past work to automatically extract material property information from literature has focused on specific properties typically using keyword search methods or regular expressions \cite{friedl2006mastering}.
However, there are few solutions in the literature that address building general-purpose capabilities for extracting material property information, i.e., for any material property. Moreover, property extraction and analysis of polymers from a large corpus of literature has also not yet been addressed. Automatically analyzing large materials science corpora has enabled many novel discoveries in recent years such as Ref. \citenum{schwalbe2019graph}, where a literature extracted data set of zeolites was used to analyze interzeolite relations. Using word embeddings trained on such corpora has also been used to predict novel materials for certain applications in inorganics as well as polymers\cite{shettyautomated, tshitoyan2019unsupervised}. 

\begin{figure}
    \centering
    \includegraphics[scale=0.15]{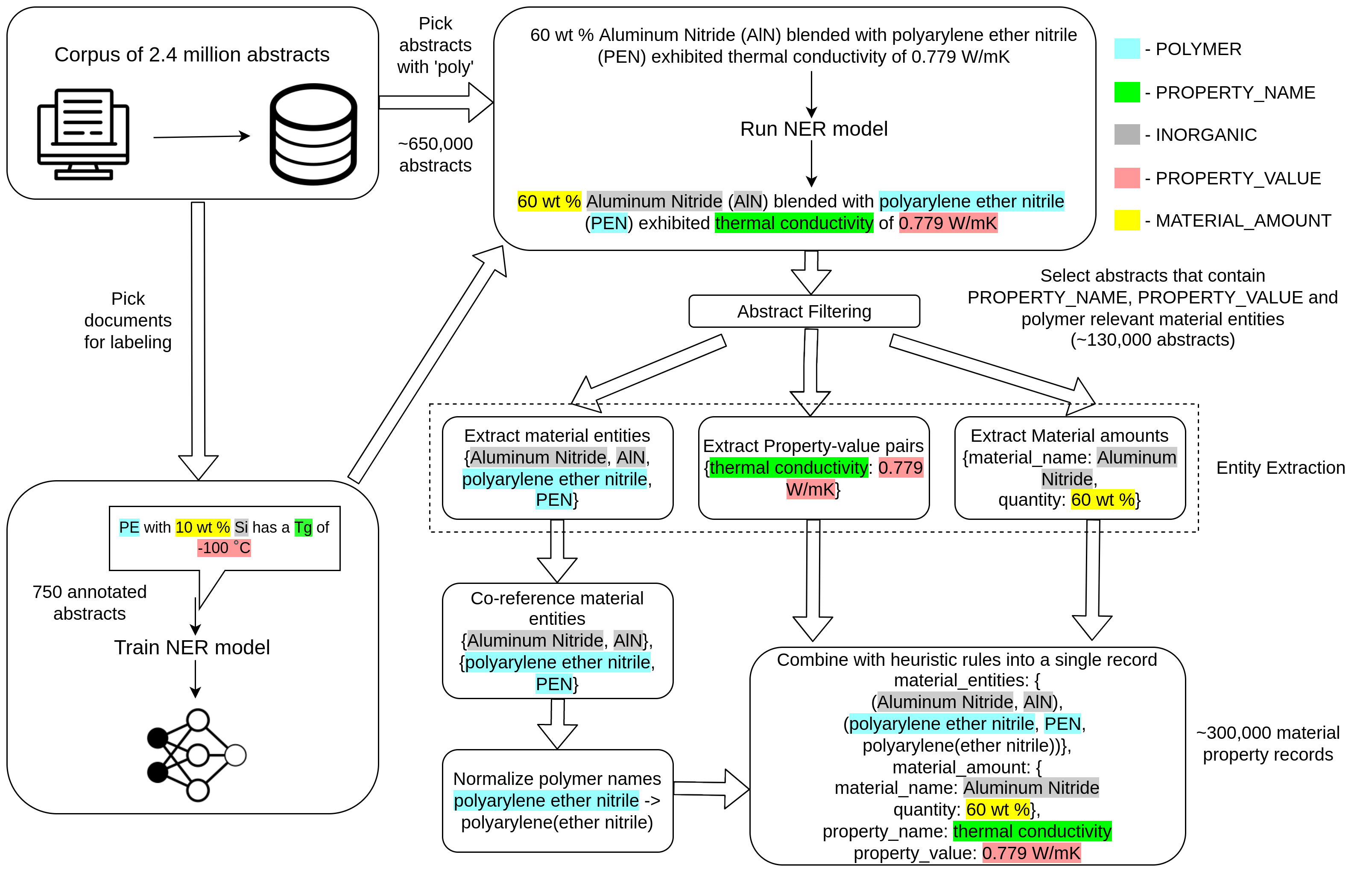}
    \caption{Pipeline used for extracting material property records from a corpus of abstracts}
    \label{fig:workflow}
\end{figure}

In this work, we built a general-purpose pipeline for extracting material property data.
Starting with a corpus of 2.4 million materials science papers described in Ref. \citenum{shettyautomated}, we selected $750$ abstracts from the polymer domain and annotated each of the abstracts using our own ontology that was designed for the purpose of extracting information from materials science literature.
Using these $750$ annotated abstracts we trained an NER model, using our MaterialsBERT language model to encode the input text into vector representations. MaterialsBERT in turn was trained by starting from PubMedBERT, another language model, and using 2.4 million materials science abstracts to continue training the model \cite{gu2020domain}.
The trained NER model was applied to polymer abstracts and heuristic rules were used to combine the predictions of the NER model and obtain material property records from all polymer relevant abstracts. This pipeline is illustrated in Figure \ref{fig:workflow}.
We restricted our focus to abstracts as associating property value pairs with their corresponding materials is a more tractable problem in abstracts.
We analyzed the data obtained using this pipeline for applications as diverse as polymer solar cells, fuel cells, and supercapacitors and showed that several known trends and phenomena in materials science can be inferred using this data.
Moreover, we trained a machine learning predictor for the glass transition temperature using automatically extracted data (Supplementary Information Section S5).

This is the first work to build a general-purpose material property data extraction pipeline, for any  material property. MaterialsBERT, the language model that powers our information extraction pipeline is released in order to enable the information extraction efforts of other materials researchers. We show that MaterialsBERT outperforms other similar BERT-based language models such as BioBERT \cite{lee2020biobert} and ChemBERT \cite{guo2021automated} on three out of five materials science NER data sets. The data extracted using this pipeline can be explored using a convenient web-based interface (\url{https://www.polymerscholar.org}) which can aid polymer researchers in locating material property information of interest to them.

\section{Results}

\subsection{Abstract annotation}
Our ontology for extracting material property information consists of 8 entity types namely POLYMER, POLYMER\_CLASS, PROPERTY\_VALUE, PROPERTY\_NAME, MONOMER, ORGANIC\_MATERIAL, INORGANIC\_MATERIAL, and MATERIAL\_AMOUNT. For a detailed description of these entity types, see Table \ref{baseline}. This ontology captures the key pieces of information commonly found in abstracts and the information we wish to utilize for downstream purposes. Note that, unlike some other studies, our ontology scheme does not annotate entities using the BIO tagging scheme, i.e., \textbf{B}eginning-\textbf{I}nside-\textbf{O}utside of the labeled entity. Instead, we opt to keep the labels simple and annotate only tokens belonging to our ontology and label all other tokens as `OTHER'. This is because, as reported in Ref. \citenum{gu2020domain}, for BERT-based sequence labeling models, the advantage offered by explicit BIO tags is negligible and IO tagging schemes suffice. % Give an example of BIO and contrast with IO
More detailed annotation guidelines are provided in the Supplementary Information Section S1.
% Add specific section
A corpus of 2.4 million materials science papers was filtered to obtain a dataset of abstracts that were polymer relevant and likely to contain the entity types of interest to us. We did so by filtering abstracts containing the string `poly' to find polymer-relevant abstracts and used regular expressions to find abstracts that contained numeric information.

Using the above-described ontology, we annotated 750 abstracts and split the abstracts into \SI{80}{\%} for training, \SI{10}{\%} for validation, and \SI{10}{\%} for testing. Prior to manual annotation, we pre-annotated the dataset using dictionaries of entities for the entity types where one was available \cite{weston2019named}. This was intended to speed up the annotation process. This dataset was annotated by three domain experts using the tool Prodigy (https://prodi.gy). Annotation was done over three rounds using a small sample of abstracts in each round. With each round, the annotation guidelines were refined and the abstracts in the previous rounds were re-annotated using the refined guidelines.

In order to assess the inter-annotator agreement between the three annotators, we use 10 common abstracts to measure Cohen's Kappa and Fleiss Kappa \cite{fleiss1971measuring} metrics. The Fleiss Kappa metric was computed to be 0.885 and the pairwise Cohen's Kappa metric to be (0.906, 0.864, 0.887) for each of the three pairs of annotators. These metrics are comparable to inter-annotator agreements reported elsewhere in the literature \cite{tabassum2020wnut} and indicate good homogeneity in the annotations.

\begin{table}[!h]
\caption{Description of each entity type in the ontology used for annotating polymer abstracts}
\begin{center}
\begin{tabular}{|c|p{8cm}|}
    \hline
    \textbf{Entity type} & \textbf{Description}\\
    \hline
    POLYMER & Material entities that are polymers \\
    \hline
    ORGANIC\_MATERIAL & Material entities that are organic but not polymers. Typically used as plasticizers or cross-linking agents \\
    \hline
    MONOMER & Material entities which are explicitly indicated as being the repeat units for a POLYMER entity \\
    \hline
    POLYMER\_CLASS & Material entities that don't refer to a specific chemical substance but are broad terms used for a class of polymers\\
    \hline
    INORGANIC\_MATERIAL & Material entities which are inorganic and are typically used as additives in a polymer formulation \\
    \hline
    MATERIAL\_AMOUNT & Entity type indicating the amount of a particular material in a material formulation \\
    \hline
    PROPERTY\_NAME & Entity type for a material property \\
    \hline
    PROPERTY\_VALUE  & Entity type including a numeric value and its unit for a material property \\
    \hline
    OTHER & Default entity type used for all tokens that do not lie in any of the above categories \\
    \hline
\end{tabular}
\end{center}
\label{baseline}
\end{table}

\subsection{NER model}
\label{sec:NER}

\begin{figure}
    \centering
    \includegraphics[scale=0.18]{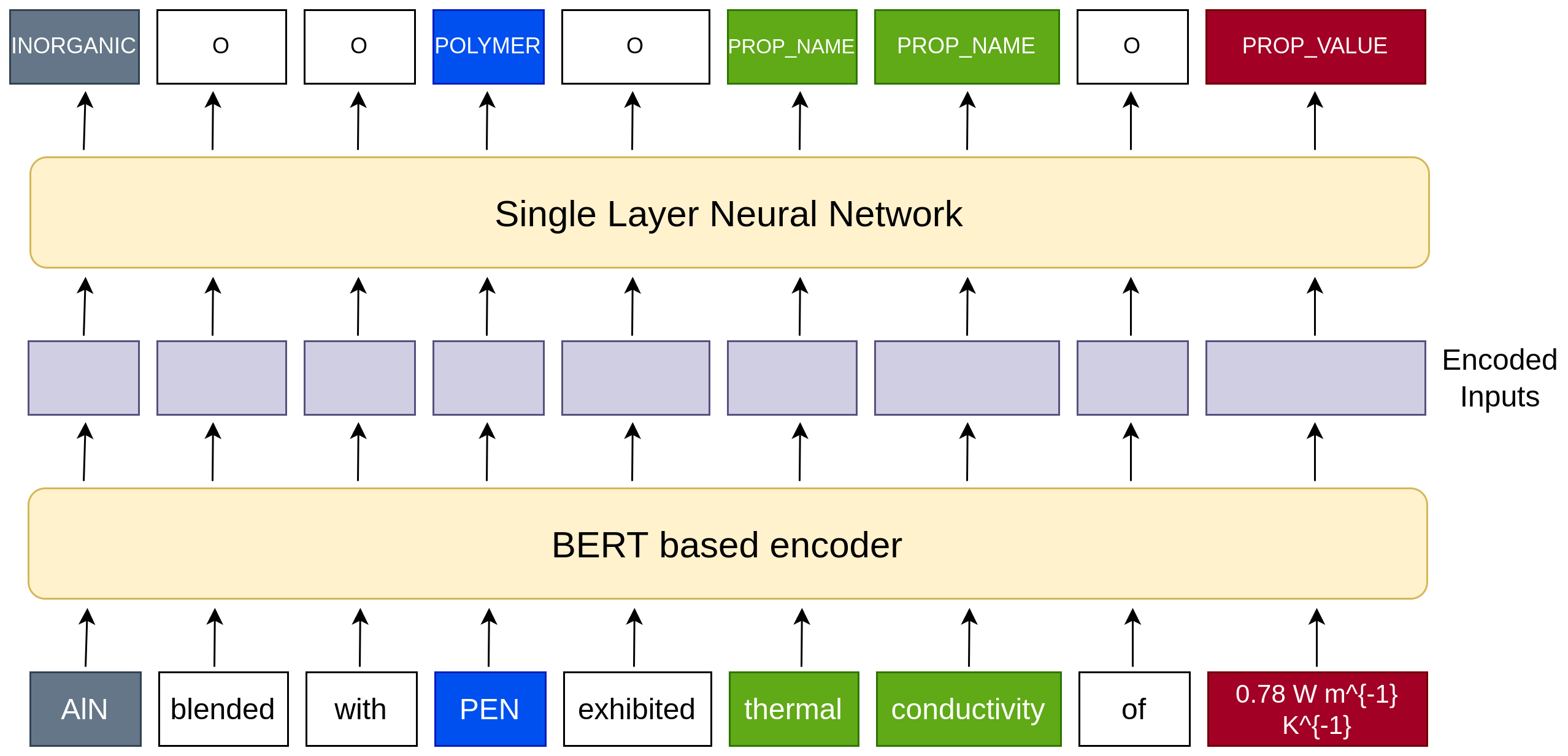}
    \caption{Model architecture used for named entity recognition: Each token in the input sequence is converted to a contextual embedding by a pre-trained transformer model which is then input to a single layer neural network. The output of the neural network is the entity type of the input token.}
    \label{fig:NER}
\end{figure}

The architecture used for training our NER model is depicted in Figure \ref{fig:NER}.
BERT and BERT-based models have become the de-facto solutions for a large number of NLP tasks\cite{devlin2018bert}. It embodies the transfer learning paradigm in which a model is trained on a large amount of unlabeled text using unsupervised objectives (not shown in the figure). The resulting BERT encoder can be used to generate token embeddings for the input text that are conditioned on all other input tokens and hence are context-aware. We used a BERT-based encoder to generate representations for tokens in the input text as shown in Figure \ref{fig:NER}. The generated representations were used as inputs to a linear layer connected to a softmax non-linearity that predicted the entity type of each token. We used a number of different encoders and compared the performance of the resulting models on our dataset of annotated polymer abstracts. We compared these models for a number of different publicly available materials science datasets as well. 

We used early stopping while training the NER model, i.e., the number of epochs of training was determined by the peak performance of the model on the validation set as evaluated after every epoch of training \cite{liang2020bond}. We also fine-tuned PubMedBERT on the abstracts of the 2.4 million papers, referred to as MaterialsBERT. During fine-tuning, we load the model and weights of PubMedBERT and continue training using the unsupervised objectives that PubMedBERT was originally trained on. The text of our corpus of abstracts forms the unlabeled input to the model. This is one of the BERT-based encoders that we test.

\subsection{Evaluation methods}

The performance of the NER model is evaluated using precision, recall and F1 score of the predicted entity tag compared to the ground truth labels. These are defined as below:

\begin{equation}
\begin{split}
    Precision &= \frac{TP}{TP+FP} \\
    Recall &= \frac{TP}{TP+FN} \\
    F1 &= \frac{2 \times Precision \times Recall}{Precision + Recall}
\end{split}
\end{equation}

\noindent where TP are the true positives, FP are the false positives and FN are the false negatives. Each of the above metrics is reported as a \% value.
We consider a predicted label to be a true positive only when the label of a complete entity is predicted correctly. For instance, for the polymer `polyvinyl ethylene', both `polyvinyl' and `ethylene' must be correctly labeled, else the entity is deemed to be predicted incorrectly.

\subsection{NER model performance}
\label{subsection:model_performance}

\begin{table}
\caption{Performance of various pre-trained BERT-based encoders on the test set of polymer abstracts. Values are reported in \%}
\begin{center}
\begin{tabular}{|c|c|c|c|}
    \hline
    \textbf{Model} & \textbf{Precision} & \textbf{Recall} & \textbf{F1}\\
    \hline
    MaterialsBERT (ours) & 62.5 & 70.6 & \textbf{66.4} \\
    \hline
    PubMedBERT & 61.4 & 70.7 & 65.8 \\
    \hline
    MatBERT & 60.9 & 70.1 & 65.2 \\
    \hline
    BioBERT & 59.2 & 66.3 & 62.6 \\
    \hline
    ChemBERT & 52.2 & 62.6 & 57.0 \\
    \hline
    BERT-base & 52.1 & 61.0 & 56.2 \\
    \hline
\end{tabular}
\end{center}
\label{ner_metrics}
\end{table}

The performance of various pre-trained BERT-based language models tested for training an NER model using our annotated data set of polymer abstracts is shown in Table \ref{ner_metrics}. We observe that MaterialsBERT, the model fine-tuned by us on 2.4 million materials science abstracts using PubMedBERT as the starting point, outperforms PubMedBERT as well as other language models used. This is in agreement with results previously reported where the fine-tuning of a transformer-based language model on a domain-specific corpus results in improved downstream task performance \cite{gu2020domain}.
Similar trends are observed across two of the four materials science data sets as reported in Table \ref{mse_datasets} and thus MaterialsBERT outperforms other pre-trained language models in three out of five materials science data sets. These NER datasets were chosen to span a range of subdomains within materials science, i.e., across organic and inorganic materials. A more detailed description of these NER datasets is provided in Supplementary Information Section S2. Note that all pre-trained encoders tested in Table \ref{ner_metrics} use the BERT-base architecture, differing in their weights and hence are comparable. MaterialsBERT outperforms PubMedBERT on all datasets except ChemDNER, which demonstrates that fine-tuning on a domain specific corpus indeed produces a performance improvement on downstream classification tasks. ChemBERT is BERT-base fine-tuned on a corpus of $\sim$400,000 organic chemistry papers and also out-performs BERT-base \cite{devlin2018bert} across the NER data sets tested. BioBERT \cite{lee2020biobert} was trained by fine-tuning BERT-base using the PubMed corpus and thus has the same vocabulary as BERT-base in contrast to PubMedBERT which has a vocabulary specific to the biomedical domain. Ref. \citenum{trewartha2022quantifying} describes the model MatBERT which was pre-trained from scratch using a corpus of 2 million materials science articles. Despite MatBERT being a model that was pre-trained from scratch, MaterialsBERT outperforms MatBERT on three out of five datasets. While the vocabulary of MatBERT and MaterialsBERT are both relevant to the domain of materials science, this performance difference can likely be attributed to the fact that PubMedBERT, the initial model for MaterialsBERT was pre-trained on a much larger corpus of text (14 million abstracts and full text). All experiments shown in Table \ref{ner_metrics} and Table \ref{mse_datasets} were performed by us. Note that we do not test BiLSTM-based architectures \cite{huang2015bidirectional} as past work has shown that BERT-based architectures typically outperform BiLSTM-based ones \cite{gu2020domain, trewartha2022quantifying, guo2021automated}. The performance on MaterialsBERT for each entity type in our ontology is described in Supplementary Information Section S3.

\begin{table*}[!h]
\caption{Performance of various BERT based encoders on the test sets of publicly available materials science NER datasets. Values are reported in \%}
\begin{center}
\renewcommand{\arraystretch}{1.4}
\scalebox{0.85}{
\begin{tabular}{|p{2.9cm}|c|c|c|c|c|c|c|c|c|c|c|c|}
    \hline
    Pre-trained encoder & \multicolumn{3}{p{3.3cm}|}{ChemDNER \cite{krallinger2015chemdner}} & \multicolumn{3}{p{3.3cm}|}{Inorganic Synthesis recipes \cite{mysore2019materials}} & \multicolumn{3}{p{3.3cm}|}{Inorganic Abstracts \cite{weston2019named}} & \multicolumn{3}{p{3.7cm}|}{ChemRxnExtractor \cite{guo2021automated}} \\
    \cline{2-13}
    % \hline
     & P & R & F1 & P & R & F1 & P & R & F1 & P & R & F1 \\
    \hline
    MaterialsBERT (ours) & 70.1 & 68.2 & 69.2 & 69.1 & 68.3 & \textbf{68.6} & 85.3 & 86.7 & 86.0 & 73.5 & 69.5 & \textbf{71.4}\\
    \hline
    PubMedBERT & 71.5 & 69.0 & \textbf{70.2} & 69.9 & 65.3 & 67.6 & 84.0 & 86.2 & 85.0 & 68.1 & 59.5 & 63.6\\
    \hline
    MatBERT & 71.7 & 66.9 & 69.2 & 68.6 & 67.7 & 68.2 & 85.6 & 86.7 & 86.2 & 67.4 & 58.0 & 62.4\\
    \hline
    BioBERT & 70.6 & 65.7 & 68.0 & 64.4 & 63.7 & 64.0 & 85.6 & 87.1 & \textbf{86.4} & 74.8 & 65.4 & 69.8\\
    \hline
    ChemBERT & 72.5 & 66.4 & 69.4 & 66.8 & 64.3 & 65.6 & 83.2 & 86.4 & 84.8 & 65.0 & 64.0 & 64.4\\
    \hline
    BERT-base & 71.2 & 65.7 & 68.4 & 62.4 & 60.3 & 61.4 & 81.0 & 81.9 & 81.4 & 57.7 & 54.9 & 56.2\\
    \hline
\end{tabular}
}
\end{center}
\label{mse_datasets}
\end{table*}

\subsection{Quantifying the extracted data}

Using our pipeline, we extracted $\sim 300,000$ material property records from $\sim 130,000$ abstracts. Out of our corpus of 2.4 million articles, $\sim 650,000$ abstracts are polymer relevant and around $\sim 130,000$ out of those contain material property data. This extraction process took 60 hours using a single Quadro 16 GB GPU. To place this number in context, PoLyInfo a comparable database of polymer property records that is publicly available has 492,645 property records as of this writing \cite{otsuka2011polyinfo}. This database was manually curated by domain experts over many years while the material property records we have extracted using automated methods took 2.5 days using only abstracts and is yet of comparable size. 
The composition of these material property records is summarized in Table \ref{tab:prop_distribution} for specific properties (grouped into a few property classes) that are utilized later on. For the general property class, we compute the number of neat polymers as the material property records corresponding to a single material of the POLYMER entity type. Blends correspond to material property records with multiple POLYMER entities while composites contain at least one material entity that is not of the POLYMER or POLYMER\_CLASS entity type. To compute the number of unique neat polymer records, we first counted all unique normalized polymer names from records that had a normalized polymer name. This accounts for the majority of polymers with multiple reported names as detailed in Ref. \citenum{shetty2021machine}. Out of the remaining neat polymer records that did not have a normalized polymer name, we then counted all unique polymer names (accounting for case variations) and added them to the number of unique normalized polymer names to arrive at the estimated number of unique polymers.
The number of extracted data points reported in Table \ref{tab:prop_distribution} is higher than the figures shown later as additional constraints are imposed in each case in order to better study this data. For the general property class, we note that elongation at break data for an estimated 413 unique neat polymers was extracted. In contrast, Ref. \citenum{palomba2014prediction} used 77 polymers to train a machine learning model. For tensile strength, an estimated 926 unique neat polymer data points are extracted while Ref. \citenum{pg} used 672 data points to train a machine learning model. Thus the amount of data extracted in the aforementioned cases is already comparable to or greater than the amount of data being utilized to train property predictors in the literature. Note that Table \ref{tab:prop_distribution} accounts for only 39207 points which is 13 \% of the total extracted material property records. More details on the extracted material property records can be found in Supplementary Information Section S4. The reader is also encouraged to explore this data further through \url{https://polymerscholar.org}

\begin{table}[!h]
\centering
\begin{tabular}{|p{3cm}|@{}c|c| }% <-- aded @{}
\hline
Property class & \begin{tabular}{p{4cm}|p{3cm}|p{3cm}|p{3cm}} Property\; & Total number of datapoints & neat polymers/ blends/ composites & Estimated number of unique neat polymers \\
                 \end{tabular} \\
\hline
General & 
    \begin{tabular}{p{4cm}|p{3cm}|p{3cm}|p{3cm}}
    \hline Molecular Weight & 9053 & 9053/-/- & 2623\\
    \hline Glass Transition Temperature & 6155 & 4612/1036/507 & 1732 \\
    \hline Electrical conductivity & 6030 & 3202/606/2222 & 1017\\
    \hline Tensile Strength & 4382 & 2679/651/1052 & 926\\
    \hline Elongation at Break & 1499 & 954/234/311 & 413\\
    \end{tabular} \\
\hline
Polymer Solar Cells & 
    \begin{tabular}{p{4cm}|p{3cm}|p{3cm}|p{3cm}} Power Conversion Efficiency \;\; & 3595 & - & - \\
    \hline Open Circuit Voltage & 1386 & - & -\\
    \hline Short Circuit Current & 1049 & - & -\\
    \hline Fill Factor & 966 & - & -\\
    \end{tabular} \\
\hline
Fuel Cells & 
    \begin{tabular}{p{4cm}|p{3cm}|p{3cm}|p{3cm}}
    \hline Proton conductivity & 1359 & - & -\\
    \hline Areal Power Density & 1235 & - & -\\
    \hline Areal Current Density & 295 & - & -\\
    \hline Methanol permeability & 174 & - & -\\
    \end{tabular} \\
\hline
Supercapacitors & 
    \begin{tabular}{p{4cm}|p{3cm}|p{3cm}|p{3cm}} Gravimetric Energy Density \; & 1131 & - & -\\
    \hline Gravimetric Power Density & 898 & - & -\\
    \hline
    \end{tabular} \\
\hline
\end{tabular}
\caption{Number of material property records extracted for several key polymer properties and figures of merit for certain applications}
\label{tab:prop_distribution}
\end{table}

\subsection{General property class}

\begin{figure}[!h]
    \centering
    \includegraphics[scale=0.28]{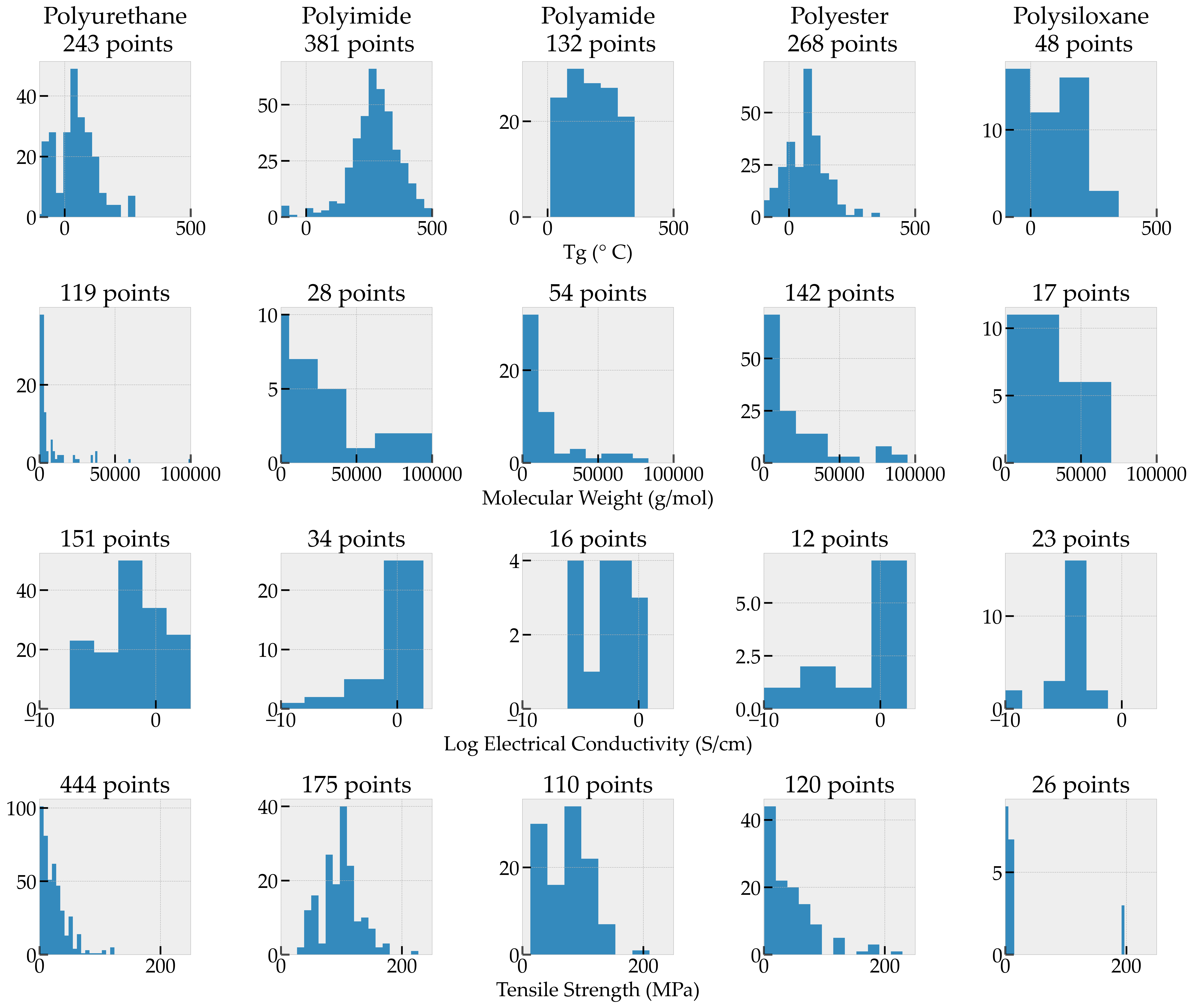}
    \caption{Material property data extracted from abstracts for material systems that contain a polymer from the polymer classes of polyurethane, polyimide, polyamide, polyester, and polysiloxane in each corresponding column. These are the most commonly reported polymer classes and the properties reported are the most commonly reported properties.}
    \label{fig:polymer_class}
\end{figure}

We now analyze the properties extracted class-by-class in order to study their qualitative trend. Figure \ref{fig:polymer_class} shows property data extracted for the five most common polymer classes in our corpus (columns) and four most commonly reported properties (rows). Polymer classes are groups of polymers that share certain chemical attributes such as functional groups. These properties fall under the general property class as described in Table \ref{tab:prop_distribution}. The data is extracted when a polymer of that polymer class is part of the formulation for which a property is reported and does not necessarily correspond to homopolymers but instead could correspond to blends or composites. The polymer class is ``inferred'' through the POLYMER\_CLASS entity type in our ontology and hence must be mentioned explicitly for the materials record to be part of this plot. Several key trends are captured in this plot. From the glass transition temperature (\Tg) row, we observe that polyamides and polyimides typically have higher \Tg than other polymer classes. Molecular weights unlike the other properties reported are not intrinsic material properties but are determined by processing parameters. The reported molecular weights are far more frequent at lower molecular weights than at higher molecular weights; mimicking a power-law distribution rather than Gaussian distribution. This is consistent with longer chains being more difficult to synthesize than shorter chains. For electrical conductivity, we find that polyimides have much lower reported values which is consistent with them being widely used as electrical insulators. Also note that polyimides have higher tensile strengths as compared to other polymer classes, which is a well-known property of polyimides \cite{zhang2018polyimide}.

\begin{figure}
    \centering
    \includegraphics[scale=0.12]{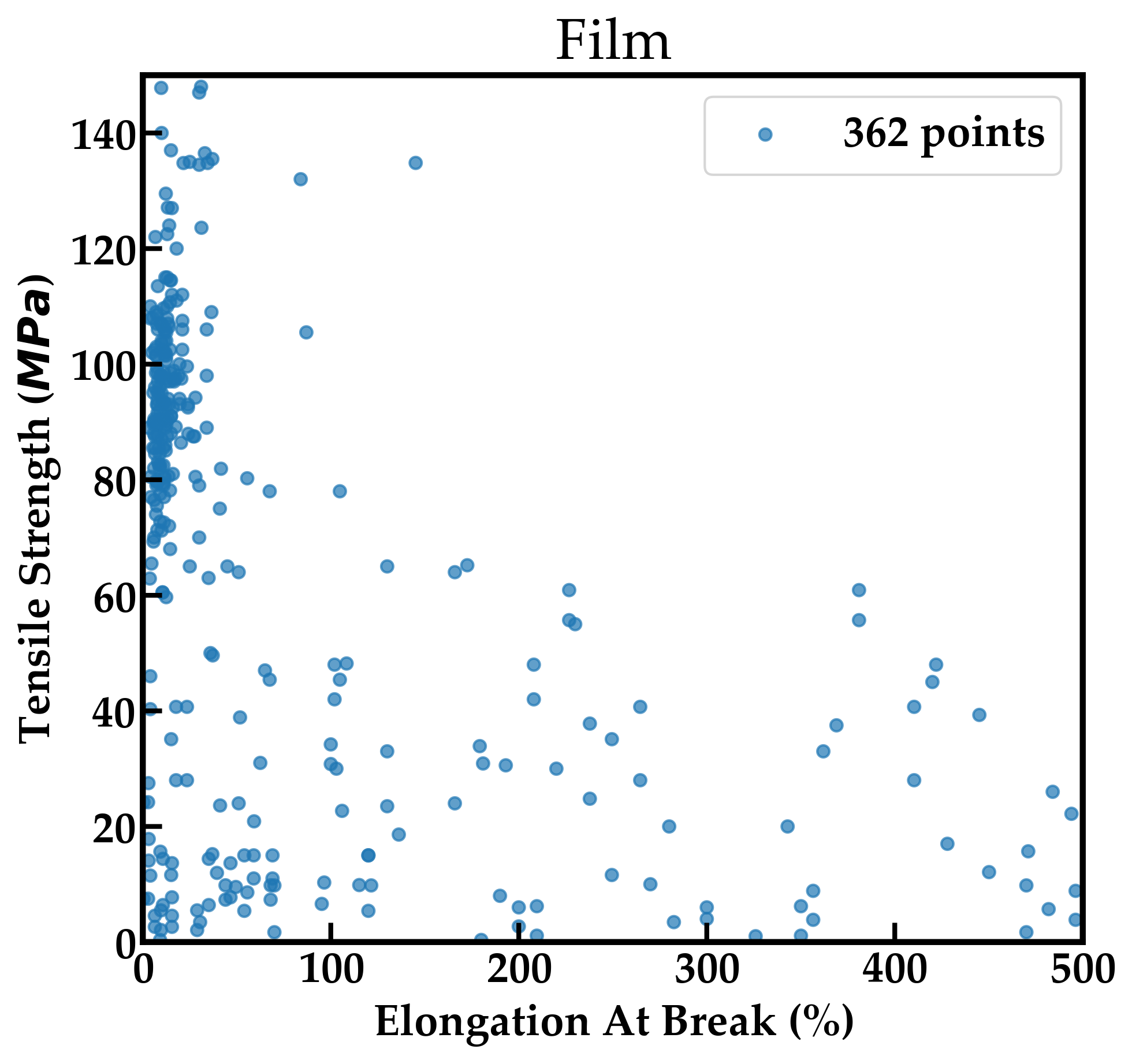}
    \caption{Tensile Strength Vs Elongation at break for films demonstrating the strength-ductility trade-off}
    \label{fig:film_str_ductility}
\end{figure}

Figure \ref{fig:film_str_ductility} shows mechanical properties measured for films which demonstrates the trade-off between elongation at break and tensile strength that is well known for material systems (often called the strength-ductility trade-off dilemma). Materials with high tensile strength tend to have a low elongation at break and conversely, materials with high elongation at break tend to have low tensile strength \cite{wang2020evading}. This known fact about the physics of material systems emerges from an amalgamation of points independently gathered from different papers. In the next section we take a closer look at pairs of properties for various devices that reveal similarly interesting trends.

\subsection{Knowledge extraction}
\label{subsection:application_analysis}

Next, we consider a few device applications and co-relations between the most important properties reported for these applications to demonstrate that non-trivial insights can be obtained by analyzing this data. We consider three device classes namely polymer solar cells, fuel cells, and supercapacitors and show that their known physics is being reproduced by NLP extracted data. We find documents specific to these applications by looking for relevant keywords in the abstract such as `polymer solar cell' or `fuel cell'. The total number of data points for key figures of merit for each of these applications is given in Table \ref{tab:prop_distribution}.

\begin{figure}
	\raggedleft
	\subfigure[]{
		\begin{minipage}{0.30\textwidth}
			\includegraphics[width=1\textwidth]{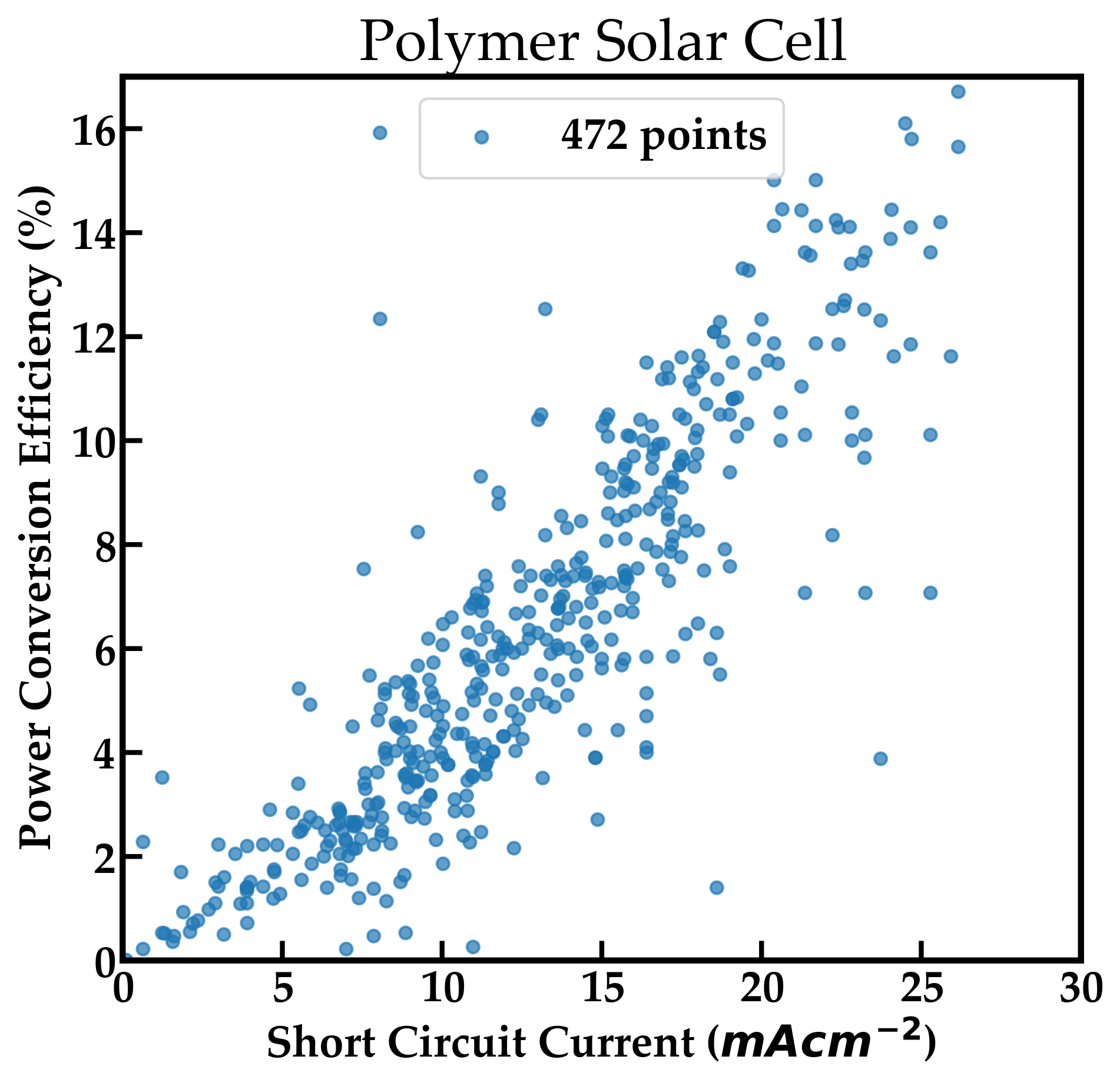}
		\end{minipage}
		\label{fig:JSC}
	}\hspace{2mm}
	\centering
	\subfigure[]{
		\begin{minipage}{0.30\textwidth}
			\includegraphics[width=1\textwidth]{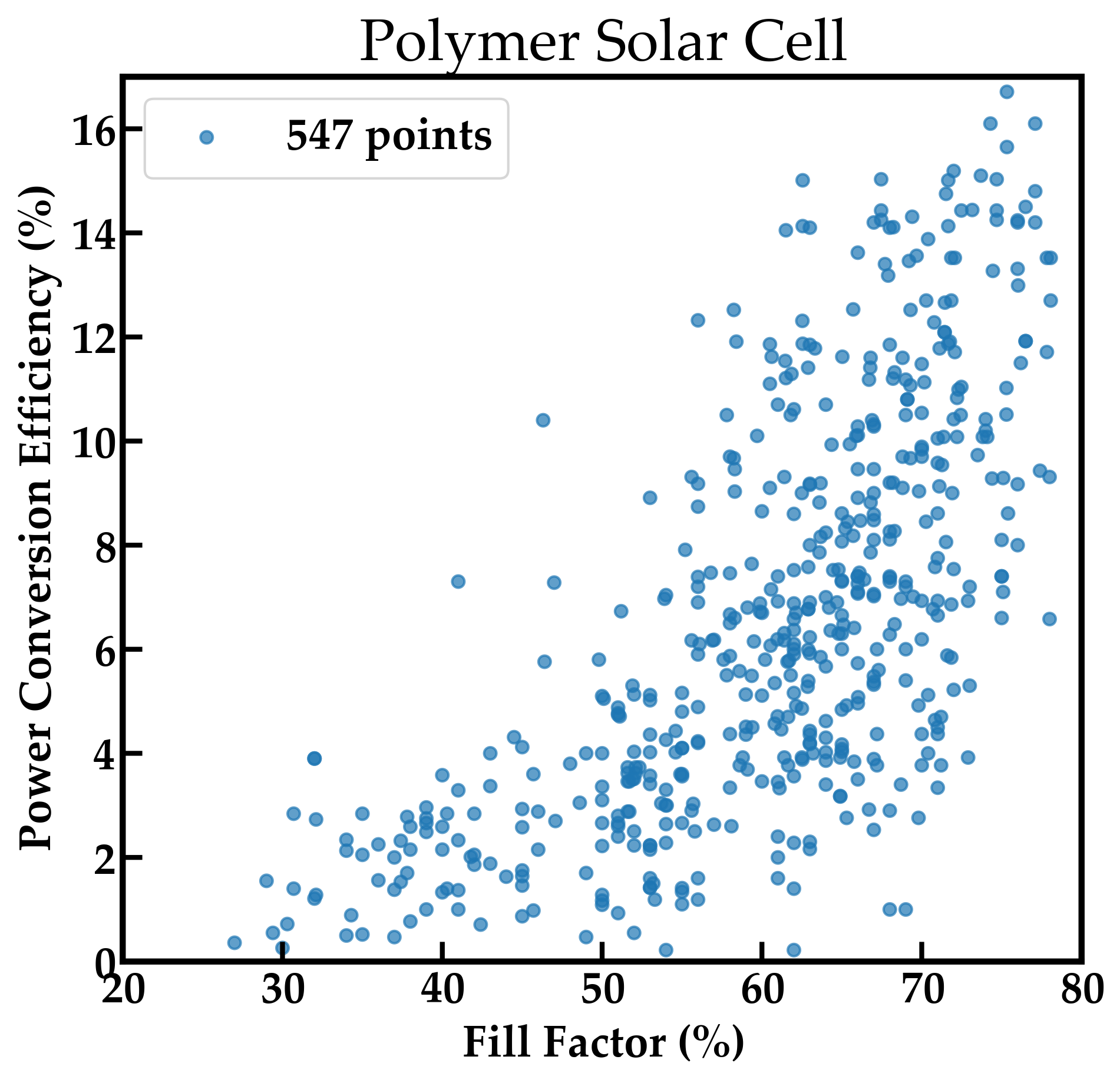}
		\end{minipage}
		\label{fig:FF}
	}
	\raggedright
	    \subfigure[]{
		\begin{minipage}{0.30\textwidth}
			\includegraphics[width=1\textwidth]{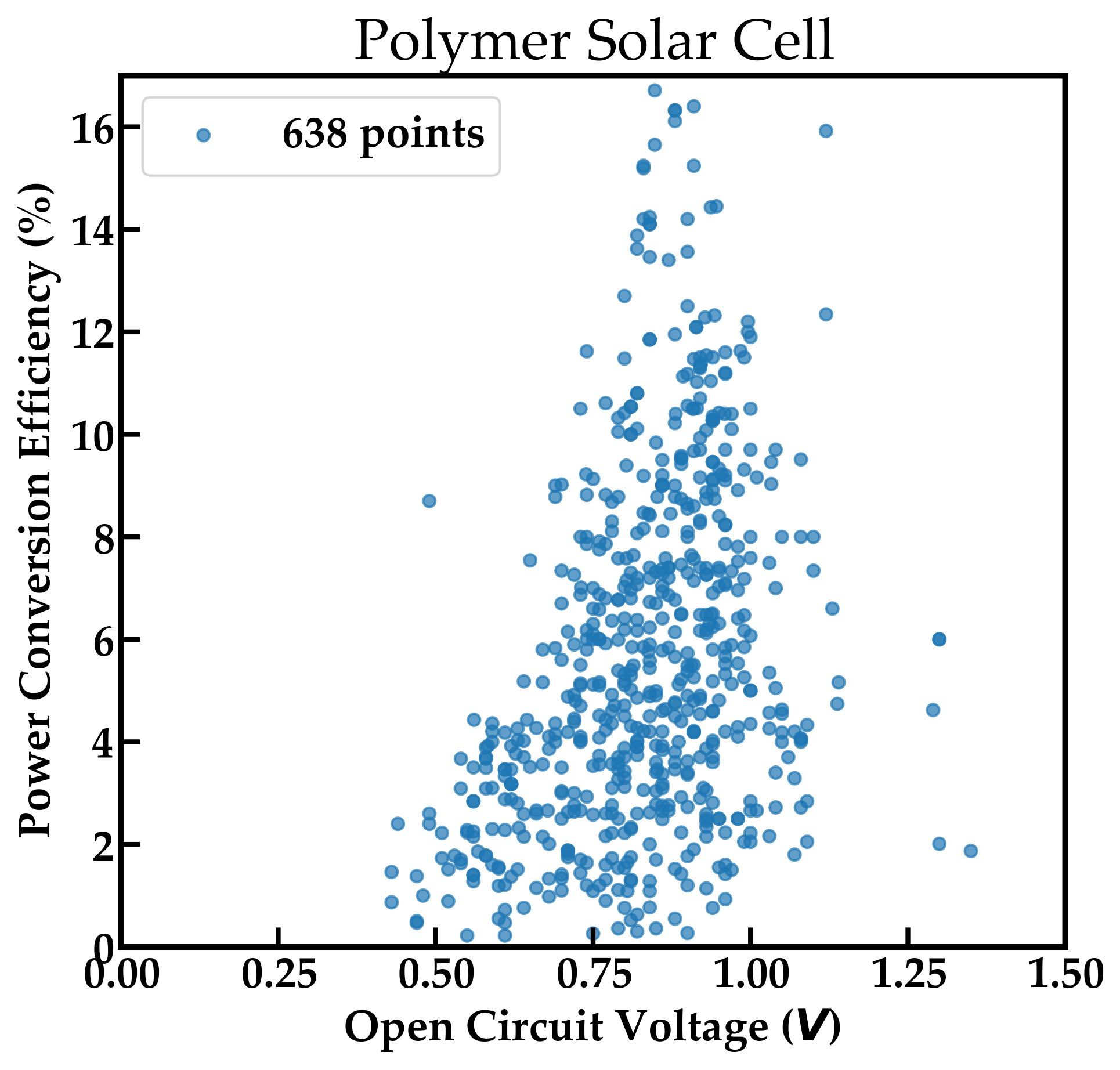}
		\end{minipage}
		\label{fig:OCV}
	}\hspace{2mm}
	
	\raggedleft
	\subfigure[]{
		\begin{minipage}{0.30\textwidth}
			\includegraphics[width=1\textwidth]{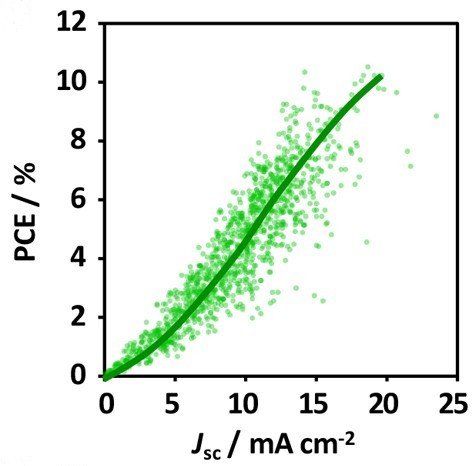}
		\end{minipage}
		\label{fig:OCV_manual}
	}\hspace{2mm}
	\centering
    \subfigure[]{
		\begin{minipage}{0.30\textwidth}
			\includegraphics[width=1\textwidth]{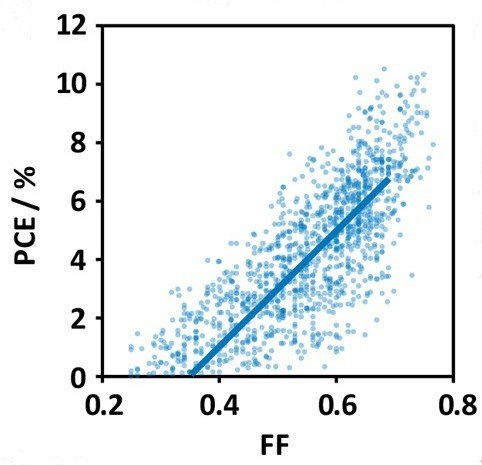}
		\end{minipage}
		\label{fig:JSC_manual}
	}
	\raggedright
	\subfigure[]{
		\begin{minipage}{0.30\textwidth}
			\includegraphics[width=1\textwidth]{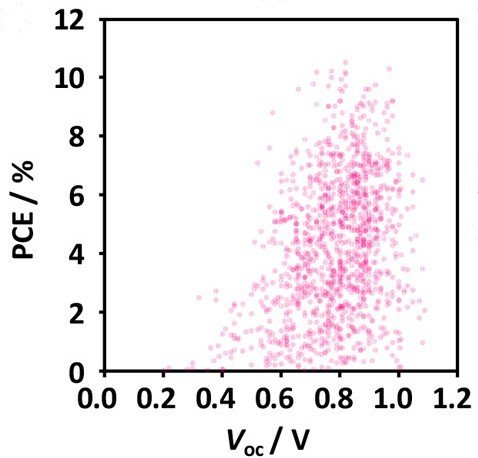}
		\end{minipage}
		\label{fig:FF_manual}
	}\hspace{2mm}
	
	\caption{Co-relations between key properties extracted automatically from literature for polymer solar cells a) Power Conversion Efficiency against short circuit current b) Power Conversion Efficiency against fill factor c) Power Conversion Efficiency against open circuit voltage. Co-relations between key properties extracted manually from literature for polymer solar cells d) Power Conversion Efficiency against short circuit current e) Power Conversion Efficiency against fill factor f) Power Conversion Efficiency against open circuit voltage. Observe that the trends in this figure match well with NLP extracted data in Figure \ref{fig:PSC}. Reproduced here with permission from Ref. \citenum{nagasawa2018computer}}
	\label{fig:PSC}
\end{figure}

\subsubsection{Polymer solar cells}
Polymer solar cells, in contrast to conventional silicon-based solar cells, have the benefit of lower processing cost but suffer from lower power conversion efficiencies. Improving their power conversion efficiency by varying the materials used in the active layer of the cell is an active area of research \cite{psc}. Figure \ref{fig:PSC}a)-c) shows the power conversion efficiency for polymer solar cells plotted against the corresponding short circuit current, fill factor and open circuit voltage for NLP extracted data while Figure \ref{fig:PSC}d)-f) shows the same pairs of properties for data extracted manually as reported in Ref. \citenum{nagasawa2018computer}. Each data point in Figure \ref{fig:PSC}a)-c) is taken from a particular paper and corresponds to a single material system. It is clear from Figure \ref{fig:OCV} that the peak power conversion efficiencies reported are around 16.71 \% which is close to the maximum known values reported in the literature \cite{zhang2022metallated} as of this writing. The open-circuit voltages (OCV) appear to be Gaussian distributed at around \SI{0.85}{V}. Figure \ref{fig:JSC} shows a linear trend between short circuit current and power conversion efficiency. It is clear that the trends observed in Figure \ref{fig:PSC}a)-c) for NLP extracted data are quite similar to the trends observed from manually curated data in Figure \ref{fig:PSC}d)-f).

\subsubsection{Fuel Cells}

\begin{figure}%\ContinuedFloat
    \raggedleft
    \subfigure[]{
		\begin{minipage}{0.30\textwidth}
			\includegraphics[width=1\textwidth]{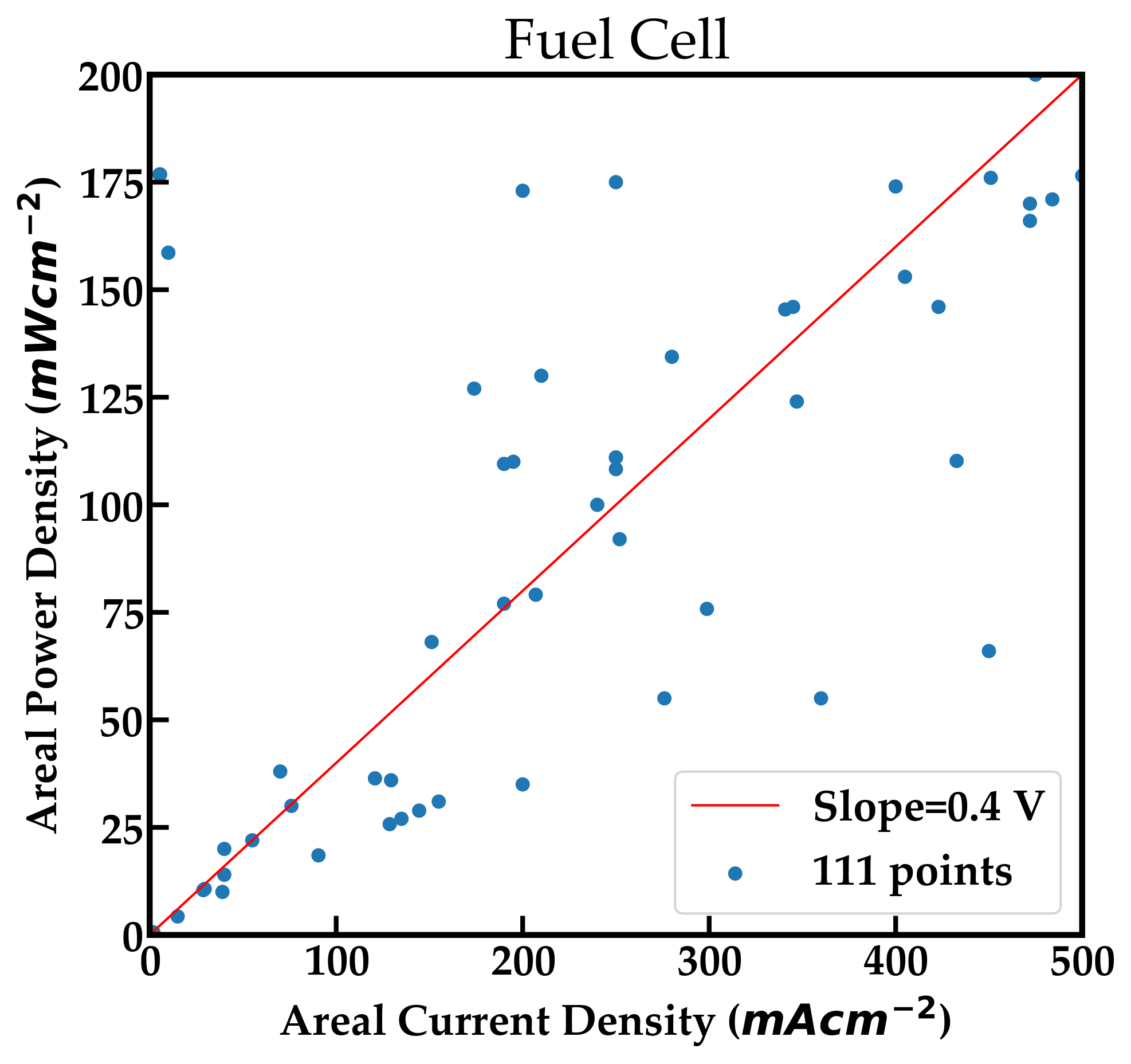}
		\end{minipage}
		\label{fig:APD_fuel_cell}
	}
	\raggedright
	\subfigure[]{
		\begin{minipage}{0.30\textwidth}
			\includegraphics[width=1\textwidth]{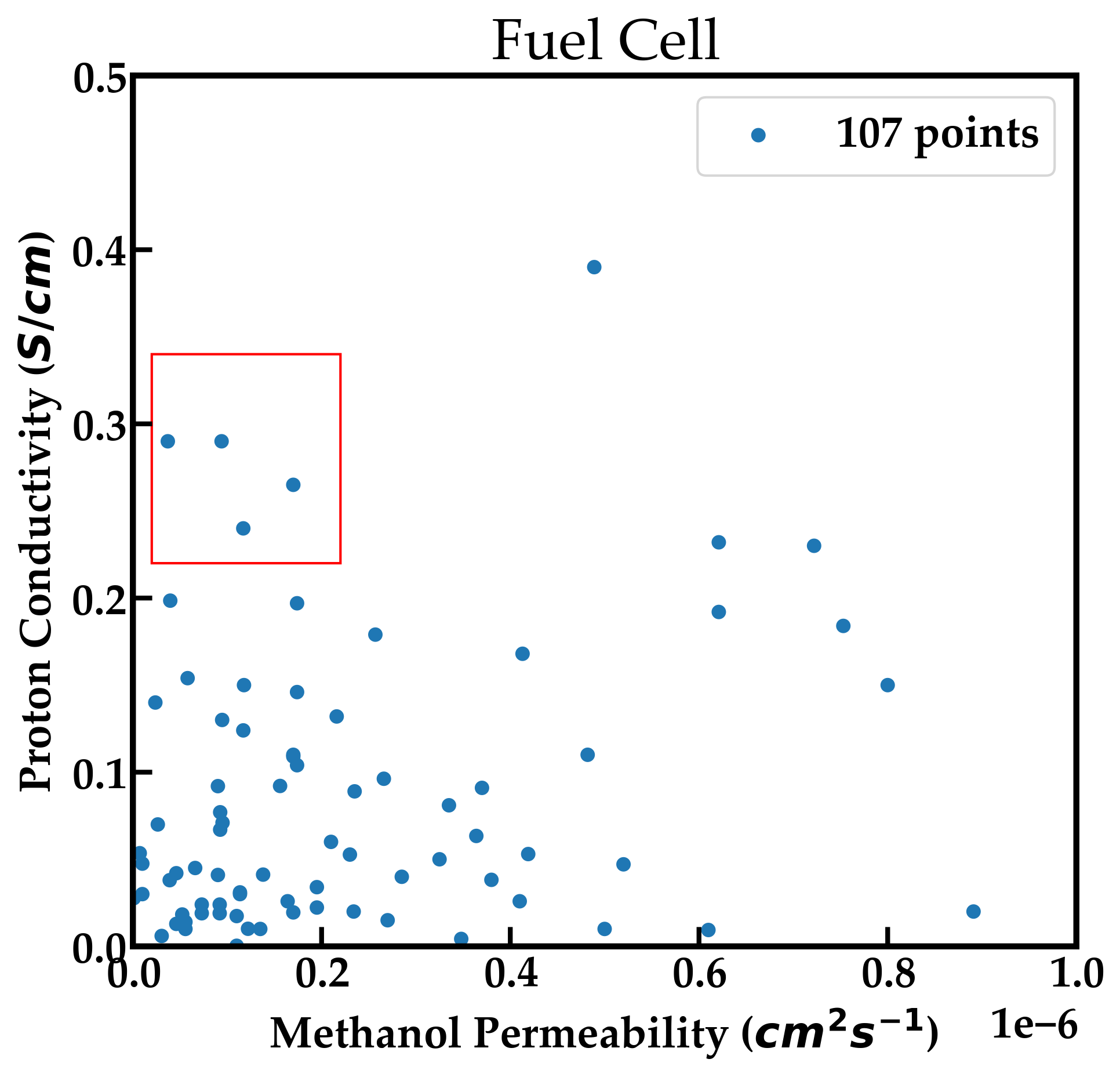}
		\end{minipage}
		\label{fig:MPD_fuel_cell}
	}\hspace{2mm}
	\raggedright
	\subfigure[]{
		\begin{minipage}{0.30\textwidth}
			\includegraphics[width=1\textwidth]{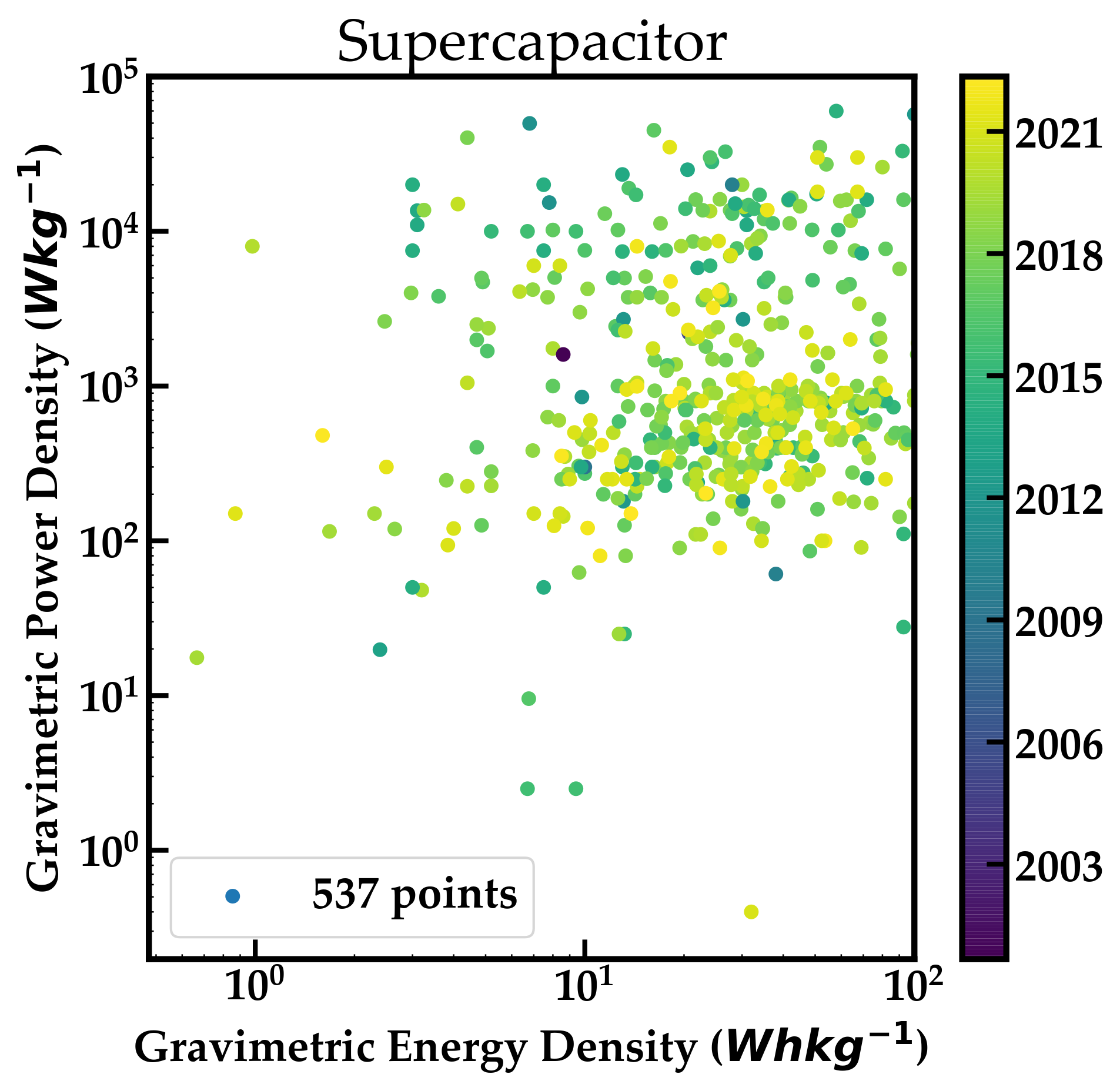}
		\end{minipage}
		\label{fig:supercapacitor}
	}\hspace{2mm}
	\raggedleft
    \subfigure[]{
		\begin{minipage}{0.30\textwidth}
			\includegraphics[width=1\textwidth]{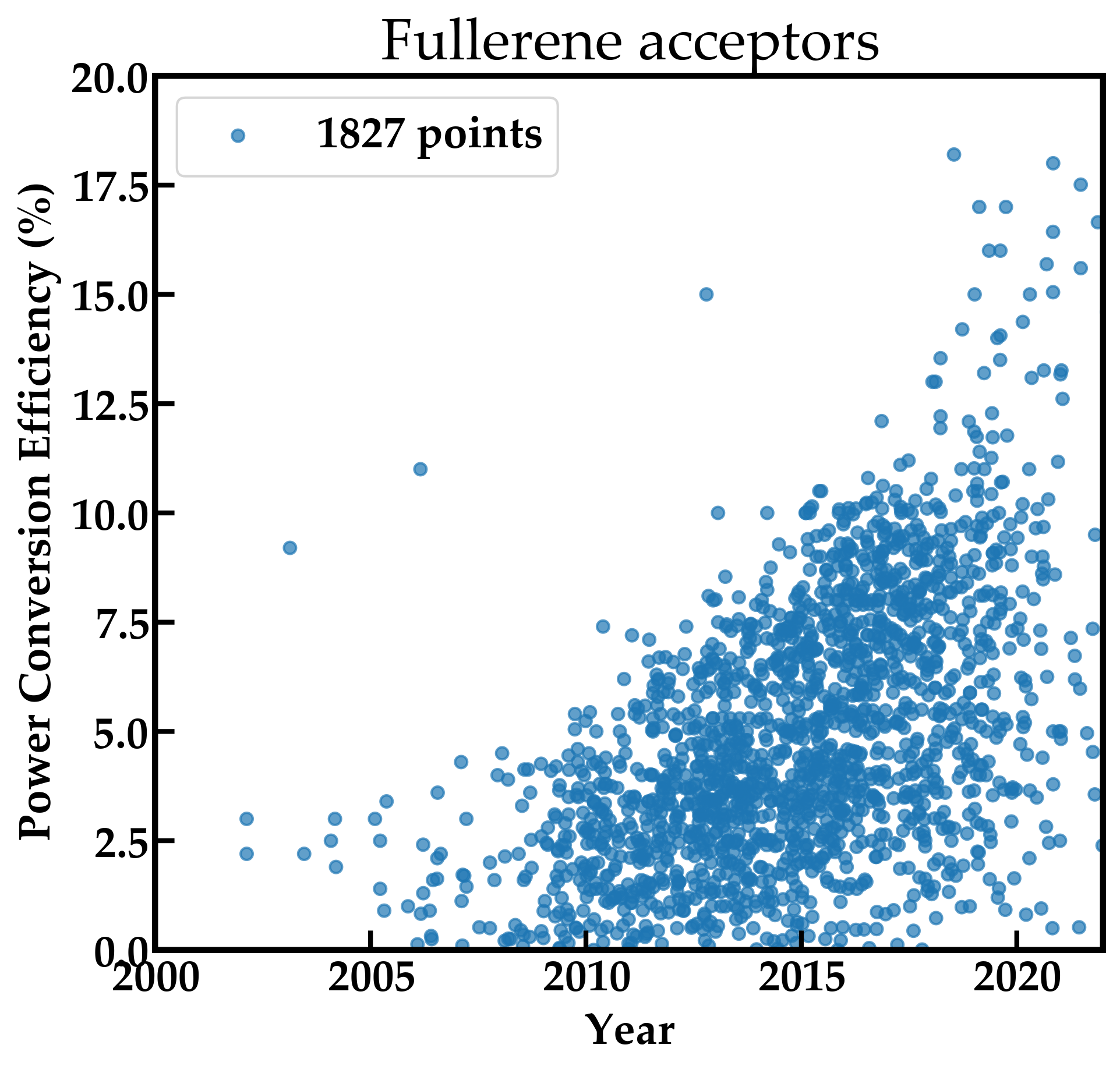}
		\end{minipage}
		\label{fig:PCE_fullerene}
	}
	\centering
	\subfigure[]{
		\begin{minipage}{0.30\textwidth}
			\includegraphics[width=1\textwidth]{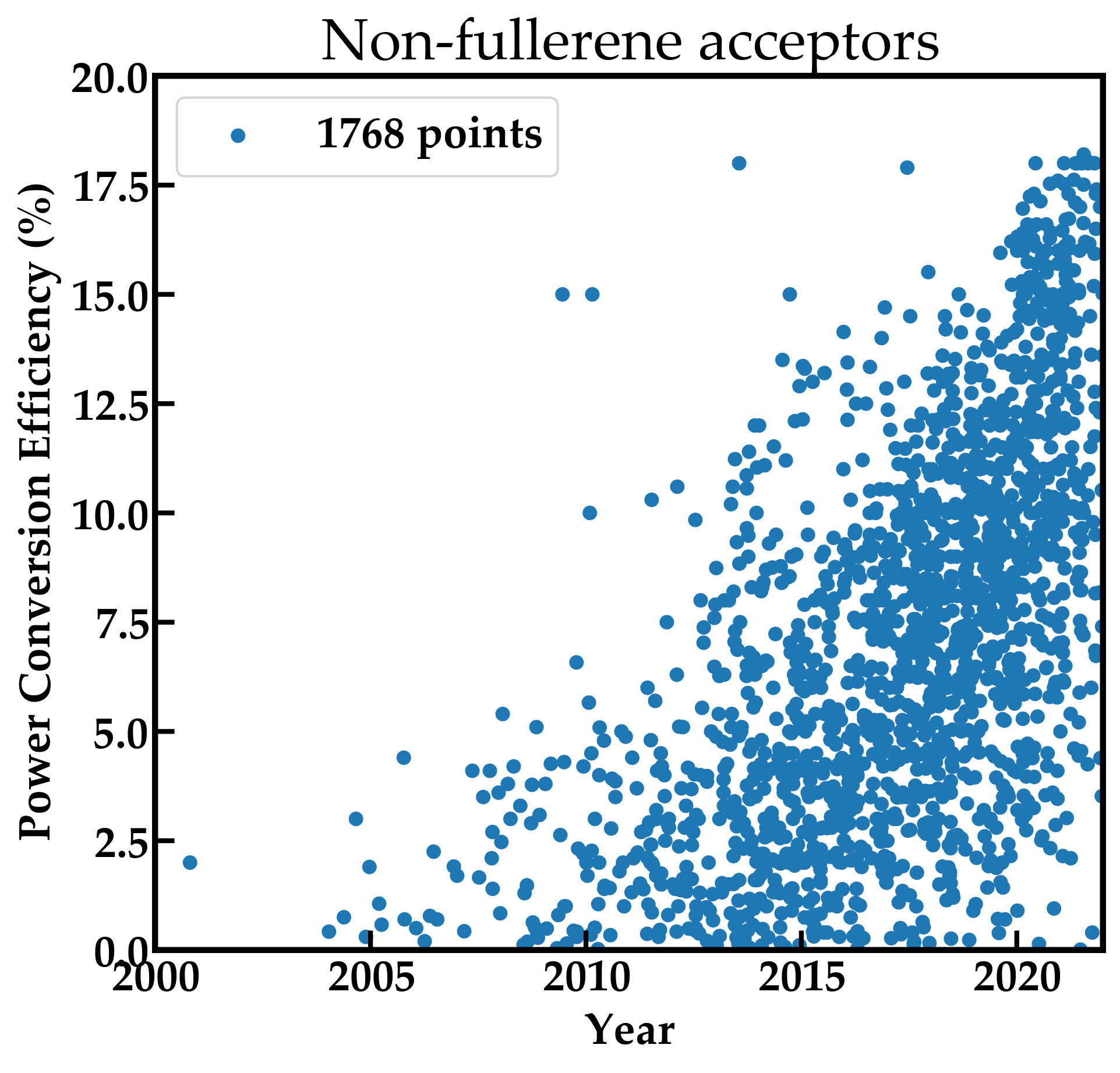}
		\end{minipage}
		\label{fig:PCE_nonfullerene}
	}\hspace{2mm}
	\raggedright
	\subfigure[]{
		\begin{minipage}{0.32\textwidth}	\includegraphics[width=1.1\textwidth]{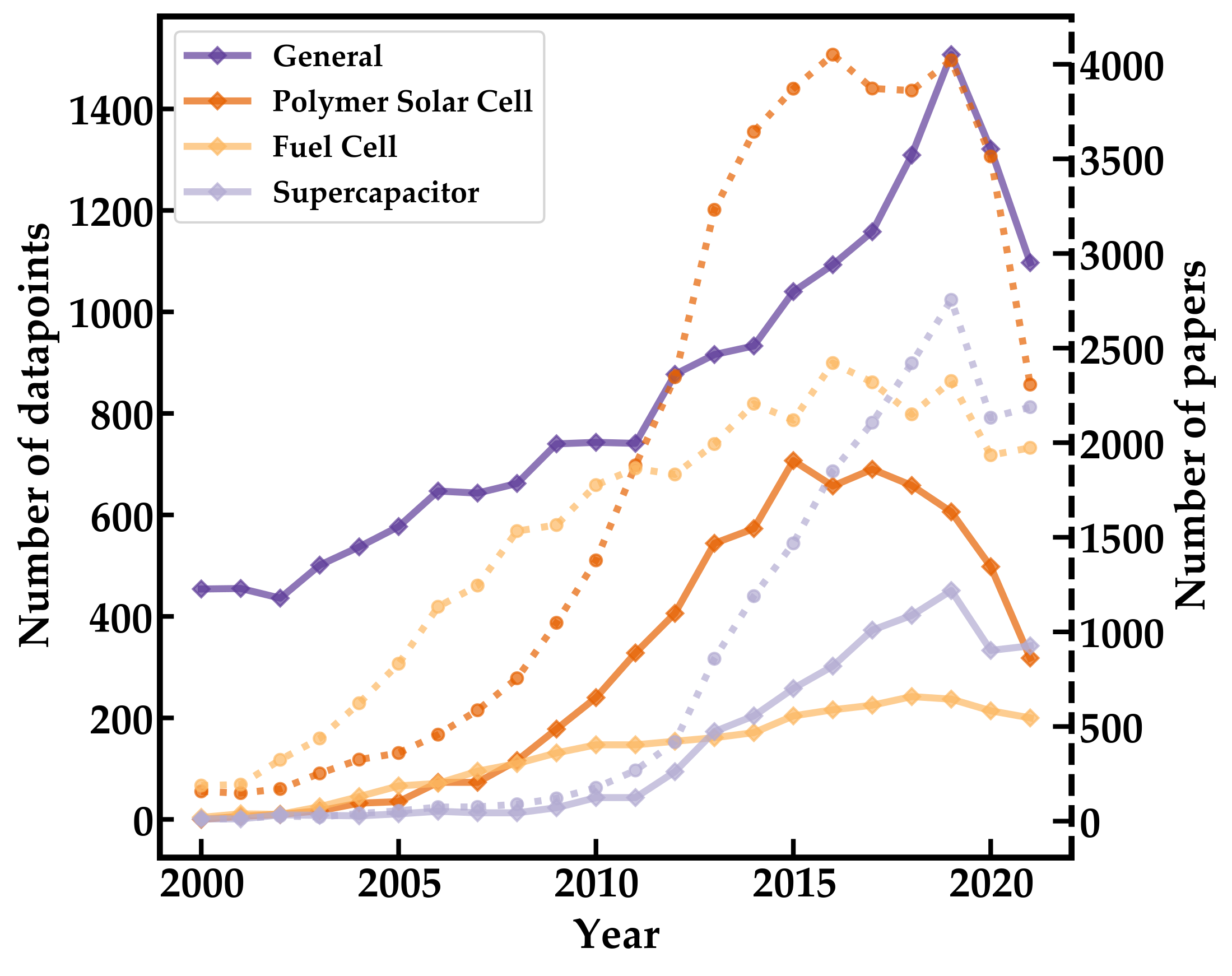}
		\end{minipage}
		\label{fig:datapoints_trend}
	}\hspace{2mm}

	\caption{Co-relations between key properties extracted automatically from literature for three different applications a) Areal current Density Vs Areal Power Density for fuel cells. the slope of the best bit line has a slope of 0.4V which is the typical operating voltage of a fuel cell b) Proton Conductivity Vs Methanol permeability for fuel cells. The red box shows the desirable region of the property space c) Up-to-date Ragone plot for supercapacitors showing energy density Vs power density. d) Power Conversion Efficiency against time for fullerene acceptors and e) Power Conversion Efficiency against time for non-fullerene acceptors f) Trend of number of datapoints extracted by our pipeline over time. The dashed lines represent the number of papers published for each of the three applications in the plot and correspond to the dashed Y-axis.}
% 	\vspace{-3mm}
	\label{fig:corelations}
% 	\vspace{-3mm}
\end{figure}

Fuel cells are devices that convert a stream of fuel such as methanol or hydrogen and oxygen to electricity. Water is one of the primary by-products of this conversion making this a clean source of energy. A polymer membrane is typically used as a separating membrane between the anode and cathodes in fuel cells \cite{abdelkareem2021environmental}. Improving the proton conductivity and thermal stability of this membrane to produce fuel cells with higher power density is an active area of research. Figure \ref{fig:APD_fuel_cell} and \ref{fig:MPD_fuel_cell} show plots for fuel cells comparing pairs of key performance metrics. The points on the power density versus current density plot (Figure \ref{fig:APD_fuel_cell}) lie along the line with a slope of \SI{0.4}{V} which is the typical operating voltage of a fuel cell under maximum current densities \cite{larminie2003fuel}. Each point in this plot corresponds to a fuel cell system extracted from the literature that typically reports variations in the polymer membrane. Figure \ref{fig:MPD_fuel_cell} illustrates yet another use-case of this capability, i.e., to find material systems lying in a desirable range of property values for the more specific case of direct methanol fuel cells. For such fuel cell membranes, low methanol permeability is desirable in order to prevent the methanol fuel from crossing the membrane and poisoning the cathode side \cite{shaari2018enhancedMP}. High proton conductivity is simultaneously desirable. The box shown in the figure illustrates the desirable region and can thus be used to easily locate promising materials systems.

\subsubsection{Trends across time}

We show that known trends across time in polymer literature are also being reproduced in our extracted data. A Ragone plot illustrates the trade-off between energy and power density for devices and supercapacitors are a class of devices that have high power density but low energy density. Figure \ref{fig:supercapacitor} illustrates the trade-off between gravimetric energy density and gravimetric power density for supercapacitors and is effectively an up-to-date version of the Ragone plot for supercapacitors \cite{catenaro2021ragone}. Historically, in most Ragone plots, the energy density of supercapacitors ranges from 1 to 10 Wh/kg \cite{shown2015conducting}. However, this is no longer true as several recent papers have demonstrated energy densities of up to 100 Wh/kg \cite{uppugalla2019polyaniline, li2019coupled, javed2019achieving}. As seen in Figure \ref{fig:supercapacitor}, the majority of points beyond an energy density of 10 Wh/kg are from the previous two years, i.e., 2020 and 2021.

Figure \ref{fig:PCE_fullerene} and Figure \ref{fig:PCE_nonfullerene} shows the evolution of power conversion efficiency of polymer solar cells for fullerene acceptors and non-fullerene acceptors. These are the two major classes of acceptors which along with a polymer donor form the active layer of a bulk heterojunction polymer solar cell. Observe that more papers with fullerene acceptors are found in earlier years with the number dropping in recent years while non-fullerene acceptor based papers have become more numerous with time. They also exhibit higher power conversion efficiencies than their fullerene counterparts in recent years. This is a known trend within the domain of polymer solar cells reported in Ref. \citenum{fu2019polymer}. It is worth noting that the authors realized this trend by studying the NLP extracted data and then looking for references to corroborate this observation.

Figure \ref{fig:datapoints_trend} shows the number of datapoints extracted by our pipeline over time for the various categories described in Table \ref{tab:prop_distribution}. Observe that the number of datapoints of the general category have grown exponentially at a rate of 6\% per year. Out of the three applications considered in Figure \ref{fig:datapoints_trend}, polymer solar cells have historically had the largest number of papers as well as datapoints although that appears to be declining over the past few years. Observe that there is a decline in the number of datapoints as well as the number of papers in 2020 and 2021. This is likely attributable to the COVID-19 pandemic \cite{ciotti2020covid} which appears to have lead to a drop in the number of experimental papers published that form the input to our pipeline \cite{gao2021potentially}.

\section{Discussion}

A natural language processing pipeline that extracts material property records from abstracts has been built and demonstrated. This however has some limitations in practice that we describe below:
\begin{enumerate}
    \item Materials property information is multi-modal and can be found in the text, tables, and figures in the body of the paper. Co-referencing material entity mentions across large spans of text and across figures and tables is a challenging problem. In addition to this, relation extraction of material entities and property value pairs occurring across sentences, are challenges that need to be addressed when extending this work from abstracts to full-text.
    \item The current ontology used consists of the most important entity types found in materials science literature. This makes it easier to combine material and property information using heuristic rules but misses other information about the material property record such as measurement methods or measurement conditions which in most cases would influence the property value.
    \item Converting polymer names to a structure (typically a SMILES string \cite{weininger1989smiles}) is also a bottleneck to training downstream models as this must be done manually. Tools that can reliably and robustly convert images of chemical structures found in the literature to SMILES string are an area of future work for the community. The SMILES string so generated can be used to generate a structural fingerprint vector of the polymer which in turn can serve as the input to a machine learning model. Expanding the scope of this pipeline to images in the body of the paper would allow training downstream property models without any additional curation for converting images to SMILES strings. Training robust property predictors in this manner would in turn allow the continuous and semi-automatic design of new materials, thus addressing a missing link in materials informatics. An example of manually converting polymer names to SMILES strings followed by the training of a property prediction model for glass transition temperature is shown in Supplementary Information Section S5.
\end{enumerate}

% \section{Summary}

The automated extraction of material property records enables researchers to search through literature with greater granularity and find materials systems in the range of interest. It also enables insights to be inferred by analyzing large amounts of literature that would not otherwise be possible. As shown in the section ``Knowledge extraction'', a diverse range of applications were analyzed using this workflow to reveal non-trivial albeit known insights. This is the first work to build a general purpose capability to extract material property records from published literature. $\sim300,000$ material property records were extracted from $\sim130,000$ polymer abstracts using this capability. Through our web interface (\url{https://polymerscholar.org}) the community can conveniently locate material property data published in abstracts. As part of this work, we also train and release MaterialsBERT, a language model that is fine-tuned on 2.4 million materials science abstracts using PubMedBERT as the starting point and obtains the best F1 score across three of five materials science NER data sets tested.

Growing the extracted material property data set further would require extending this capability to the body of the paper. This would require more robust methods to associate the entities extracted using named entity recognition. A few steps also remain in order to utilize the extracted data to produce trained machine learning property prediction models. The biggest bottleneck in the case of organic materials is obtaining SMILES strings for material entities which can then be used to generate structural fingerprints for downstream machine learning models. There is also a wealth of additional information such as processing conditions or measurement conditions that are not captured in our ontology. Addressing these bottlenecks would enable automatic and continuous updates of materials databases that can seamlessly power downstream property predictor machine learning models.

\section{Methods}

\subsection{Corpus of papers}
We have created a corpus of $\sim 2.4$ million journal articles from the materials science domain. The papers were downloaded from the APIs and websites from publishers such as Elsevier, Wiley, Royal Society of Chemistry, American Chemical Society, Springer Nature, Taylor \& Francis, and the American Institute of Physics. 
The corpus used in this work is an expanded version of the corpus described previously in Ref. \citenum{shettyautomated}. 750 abstracts of this corpus were annotated and used to train an NER model. Furthermore, the trained NER model along with heuristic rules is used to extract material property records from the abstracts of the full corpus.

\subsection{Preprocessing of documents}
Because the documents in our corpus are HTML formatted, we stripped all HTML tags to parse the plain text. Moreover, we replaced HTML superscripts and subscripts ($<$sup$>$ and $<$sub$>$) with plain text using the LaTeX convention of \^\,\{\} and \_\,\{\}, respectively. This is important in order to extract units of quantities as well as property values reported in scientific notation. Property values recorded in this notation were converted back to floating-point numbers downstream when the numeric value was to be recovered. We also mapped characters such as spaces or special characters that have multiple Unicode representations but have a similar appearance by creating a custom mapping.
% May or may not add anything here
For tokenization, i.e., breaking up text into units known as tokens which are used for downstream processing, we used wordpiece tokenization which is the standard tokenization scheme used with BERT and BERT-based models \cite{devlin2018bert, wu2016google}.
For instance `The molecular weight of the result \#\#ant P \#\#LL \#\#A - rich polymer was enhanced .' is what a sentence would look like post-tokenization. The word `resultant' and the polymer `PLLA' have been broken into sub-words tokens. This is necessary in order to tokenize arbitrary text inputs using a fixed-sized vocabulary as a larger vocabulary would increase the size of the model. Starting with a set of characters (alphabets, numbers etc), certain combinations of characters are iteratively merged and added to the vocabulary till the vocabulary reaches a certain fixed size \cite{song2020linear}. The characters to be merged are selected based on combinations that maximize the likelihood of the input text. This typically breaks up words into meaningful subunits such as `resultant' being separated into `result' and `\#\#ant' which reduces the size of the vocabulary. This does not always happen though as seen with the example of `PLLA'.

\subsection{Training MaterialsBERT}
 
BERT-base, the original BERT model, was trained using an unlabeled corpus that included English Wikipedia and the Books Corpus \cite{zhu2015aligning}. The training objectives included using the masked language modeling task, which masks a random subset of the input text and asks the language model to predict it and the next sentence prediction task, which determines for a given sentence pair whether one sentence follows the other in the training data \cite{devlin2018bert}. The vocabulary of the tokenizer was fixed at 30,000 tokens. It is known that a domain-specific BERT encoder improves performance on downstream NLP tasks for that domain because the vocabulary used for tokenization is more representative of the application of interest and because the unlabeled text is also closer to the domain of interest resulting in "better" contextual embeddings \cite{gu2020domain}. 
BERT-base was pre-trained from scratch using a general English language corpus \cite{lee2020biobert}.

Even though computationally expensive, pre-training NLP models from scratch has the advantage of creating a model with a vocabulary that is customized for the domain of interest. To give an idea of how resource-intensive this can be, note that RoBERTa, a similarly pre-trained encoder used the computing power of 1024 V100 GPUs for one day \cite{liu2019roberta}. As this is not a viable route for us, we fine-tuned a model starting from previous checkpoints. The vocabulary used while fine-tuning a model in contrast remains the same as the underlying model which is a compromise we must accept. We used PubMedBERT as our starting point and fine-tuned it using 2.4 million materials science abstracts \cite{gu2020domain}. The PubMedBERT model used here is itself pre-trained from scratch using the PubMed corpus (14 million abstracts from PubMed as well as full text articles from PubMedCentral), using the BERT-base architecture. We picked PubMedBERT as our starting point as its vocabulary is specific to the biomedical domain which overlaps with materials science as material entities are frequently mentioned in biomedical papers. During fine-tuning, the model weights of the checkpoint to be used are loaded and training is continued using the same training objectives as the model that is fine-tuned but using the unlabeled text from the fine-tuning corpus as the input. The hyperparameters used during fine-tuning were identical to those used to train PubMedBERT. We used the "Transformers" library for fine-tuning PubMedBERT \cite{wolf-etal-2020-transformers}. A similar strategy is employed in ChemBERT \cite{guo2021automated}, ClinicalBert \cite{huang2019clinicalbert}, and FinBERT \cite{araci2019finbert}. We fine-tuned PubMedBERT for 3 epochs which took 90 hours on four RTX6000 16 GB GPU's to obtain MaterialsBERT.

\subsection{Material property records extraction}

The trained NER model is one component of a system that is used to extract material property records. Each component is explained below:

\begin{enumerate}
    \item \textbf{Train NER model}: A subset of our corpus of papers is selected and annotated with a given ontology to train an NER model (described in the Section ``NER model''). This model is used to generate entity labels for abstracts in the corpus.
    
    \item \textbf{Pick documents with  `poly'}: The corpus of abstracts is down-selected by searching for the string `poly' in the abstract as a proxy for polymer relevant documents.
    
    \item \textbf{Run NER model}: The NER model previously trained is used for predicting entity labels on each polymer relevant document obtained from the previous step.
    
    \item \textbf{Abstract filtering}: As not all polymer abstracts contain material property information, the output of the NER is used as a heuristic to filter out those that do. Only abstracts with specific material entities, i.e., POLYMER, POLYMER\_FAMILY and MONOMER as well as the PROPERTY\_NAME and PROPERTY\_VALUE tags are allowed through this stage. This acts as a second filter to locate polymer relevant documents.
    
    \item \textbf{Entity extraction}: The material entities, (PROPERTY\_NAME, PROPERTY\_VALUE) and MATERIAL\_AMOUNT entities are extracted and processed separately.
    
    \item \textbf{Co-reference material entities}: This step is applied to co-reference all mentions of the same material entity. A common example of this is when a material is mentioned next to its abbreviation. We used the abbreviation detection system in ChemDataExtractor \cite{chemdataextractor} to find material entity abbreviation pairs. In addition, we co-reference material entities that are within a Levenshtein distance \cite{levenshtein} of one. 
    Co-referencing is a tractable problem in abstracts compared to full-text papers as there are no long-range dependencies and typically no anaphora resolution is required \cite{mitkov2014anaphora}.
    
    \item \textbf{Normalizing polymer names}: Polymers can have several different variations in names referring to the same chemical entity. In this step, we normalize these variations to the most commonly occurring name for that particular polymer. For instance, `poly(ethylene)' and `poly-ethylene' occurring in different abstracts are both normalized to `polyethylene'. This is done using a dictionary lookup on a dataset of polymer name clusters that were normalized using the workflow described in Ref. \citenum{shetty2021machine}. Note that we do not normalize all polymer names but only the ones which are included in our dictionary. In practice, this includes most commonly occurring polymers that have multiple names in the literature.
    
    \item \textbf{Extract Property Value pairs}: The PROPERTY\_NAME and PROPERTY\_VALUE tag are associated by co-occurrence within a context window. The numeric value of the property is separated from the units using regular expressions and all parsed property values are converted to a standard set of units. The unit used is the most commonly reported unit for that particular property. Any errors reported with the numeric value are also parsed using regular expressions.
    
    \item \textbf{Extract Material amounts}: Entities with the MATERIAL\_AMOUNT tag are extracted and the closest material entity within a context window is associated with it. 
    
    \item \textbf{Relation extraction}: In order to obtain a material property record, it is necessary to associate the material entities and the property value pair that correspond to a single record. This problem has been addressed in the literature using supervised methods \cite{wang2016relation, zhou2020nero}. However, the annotation process for relation labeling is time-intensive and hence we employ heuristics in this work to obtain relations between entities. To associate material entities with property value pairs, we associate the closest material entity tagged in the same sentence as the property value pair. If no such material entity is found then all the material entities mentioned in the abstract are associated with the property value pair. This is because, most commonly, an abstract mentions a major material system reported in the paper and reports its measured property values. This step is reasonable in abstracts, which report this information compactly, using heuristic rules. In contrast, the body of the paper would require coreferencing with information found in tables and figures to extract material property records.
    
\end{enumerate}

% Points to include
% Polymer domain sprawling, nascent polymer informatics ecosystem,
% Do what and how
% Benefits from a unique listing of names
% List out all achievements of this work and end crisply


\section{Acknowledgements}
This work was supported by the Office of Naval Research through grants N00014-19-1-2103 and N00014-20-1-2175.
% \\
\section{Competing Interests} 
The authors declare no competing interests.
% \\
% \section{Author contributions}
% Conceptualization: PS, CZ, RR, 
% Methodology: PS,
% Investigation: PS,
% Data Curation: PS, ACR, CK, SG, LPP, LH,
% Visualization: PS,
% Supervision: CZ, RR,
% Writing—original draft: PS,
% Writing—review \& editing: PS, CK, CZ, RR
% \\
\section{Data and model availability}
% The pre-trained language model MaterialsBERT is available in the HuggingFace model zoo at \url{https://huggingface.co/pranav-s/MaterialsBERT}. 
The material property data mentioned in this paper can be explored through \url{https://www.polymerscholar.org}

\bibliography{main}

\providecommand{\latin}[1]{#1}
\makeatletter
\providecommand{\doi}
  {\begingroup\let\do\@makeother\dospecials
  \catcode`\{=1 \catcode`\}=2 \doi@aux}
\providecommand{\doi@aux}[1]{\endgroup\texttt{#1}}
\makeatother
\providecommand*\mcitethebibliography{\thebibliography}
\csname @ifundefined\endcsname{endmcitethebibliography}
  {\let\endmcitethebibliography\endthebibliography}{}
\begin{mcitethebibliography}{59}
\providecommand*\natexlab[1]{#1}
\providecommand*\mciteSetBstSublistMode[1]{}
\providecommand*\mciteSetBstMaxWidthForm[2]{}
\providecommand*\mciteBstWouldAddEndPuncttrue
  {\def\EndOfBibitem{\unskip.}}
\providecommand*\mciteBstWouldAddEndPunctfalse
  {\let\EndOfBibitem\relax}
\providecommand*\mciteSetBstMidEndSepPunct[3]{}
\providecommand*\mciteSetBstSublistLabelBeginEnd[3]{}
\providecommand*\EndOfBibitem{}
\mciteSetBstSublistMode{f}
\mciteSetBstMaxWidthForm{subitem}{(\alph{mcitesubitemcount})}
\mciteSetBstSublistLabelBeginEnd
  {\mcitemaxwidthsubitemform\space}
  {\relax}
  {\relax}

\bibitem[Devlin \latin{et~al.}(2018)Devlin, Chang, Lee, and
  Toutanova]{devlin2018bert}
Devlin,~J.; Chang,~M.-W.; Lee,~K.; Toutanova,~K. Bert: Pre-training of deep
  bidirectional transformers for language understanding. \emph{arXiv preprint
  arXiv:1810.04805} \textbf{2018}, \relax
\mciteBstWouldAddEndPunctfalse
\mciteSetBstMidEndSepPunct{\mcitedefaultmidpunct}
{}{\mcitedefaultseppunct}\relax
\EndOfBibitem
\bibitem[Vaswani \latin{et~al.}(2017)Vaswani, Shazeer, Parmar, Uszkoreit,
  Jones, Gomez, Kaiser, and Polosukhin]{vaswani2017attention}
Vaswani,~A.; Shazeer,~N.; Parmar,~N.; Uszkoreit,~J.; Jones,~L.; Gomez,~A.~N.;
  Kaiser,~{\L}.; Polosukhin,~I. Attention is all you need. Advances in neural
  information processing systems. 2017; pp 5998--6008\relax
\mciteBstWouldAddEndPuncttrue
\mciteSetBstMidEndSepPunct{\mcitedefaultmidpunct}
{\mcitedefaultendpunct}{\mcitedefaultseppunct}\relax
\EndOfBibitem
\bibitem[Swain and Cole(2016)Swain, and Cole]{chemdataextractor}
Swain,~M.~C.; Cole,~J.~M. ChemDataExtractor: a toolkit for automated extraction
  of chemical information from the scientific literature. \emph{Journal of
  chemical information and modeling} \textbf{2016}, \emph{56}, 1894--1904\relax
\mciteBstWouldAddEndPuncttrue
\mciteSetBstMidEndSepPunct{\mcitedefaultmidpunct}
{\mcitedefaultendpunct}{\mcitedefaultseppunct}\relax
\EndOfBibitem
\bibitem[Rockt{\"a}schel \latin{et~al.}(2012)Rockt{\"a}schel, Weidlich, and
  Leser]{chemspot}
Rockt{\"a}schel,~T.; Weidlich,~M.; Leser,~U. ChemSpot: a hybrid system for
  chemical named entity recognition. \emph{Bioinformatics} \textbf{2012},
  \emph{28}, 1633--1640\relax
\mciteBstWouldAddEndPuncttrue
\mciteSetBstMidEndSepPunct{\mcitedefaultmidpunct}
{\mcitedefaultendpunct}{\mcitedefaultseppunct}\relax
\EndOfBibitem
\bibitem[Hawizy \latin{et~al.}(2011)Hawizy, Jessop, Adams, and
  Murray-Rust]{chemicaltagger}
Hawizy,~L.; Jessop,~D.~M.; Adams,~N.; Murray-Rust,~P. ChemicalTagger: A tool
  for semantic text-mining in chemistry. \emph{Journal of cheminformatics}
  \textbf{2011}, \emph{3}, 17\relax
\mciteBstWouldAddEndPuncttrue
\mciteSetBstMidEndSepPunct{\mcitedefaultmidpunct}
{\mcitedefaultendpunct}{\mcitedefaultseppunct}\relax
\EndOfBibitem
\bibitem[Court and Cole(2018)Court, and Cole]{court2018auto}
Court,~C.~J.; Cole,~J.~M. Auto-generated materials database of Curie and
  N{\'e}el temperatures via semi-supervised relationship extraction.
  \emph{Scientific data} \textbf{2018}, \emph{5}, 1--12\relax
\mciteBstWouldAddEndPuncttrue
\mciteSetBstMidEndSepPunct{\mcitedefaultmidpunct}
{\mcitedefaultendpunct}{\mcitedefaultseppunct}\relax
\EndOfBibitem
\bibitem[Court \latin{et~al.}(2021)Court, Jain, and Cole]{court2021inverse}
Court,~C.~J.; Jain,~A.; Cole,~J.~M. Inverse Design of Materials That Exhibit
  the Magnetocaloric Effect by Text-Mining of the Scientific Literature and
  Generative Deep Learning. \emph{Chemistry of Materials} \textbf{2021},
  \emph{33}, 7217--7231\relax
\mciteBstWouldAddEndPuncttrue
\mciteSetBstMidEndSepPunct{\mcitedefaultmidpunct}
{\mcitedefaultendpunct}{\mcitedefaultseppunct}\relax
\EndOfBibitem
\bibitem[Tchoua \latin{et~al.}(2016)Tchoua, Qin, Audus, Chard, Foster, and
  de~Pablo]{tchoua2016blending}
Tchoua,~R.~B.; Qin,~J.; Audus,~D.~J.; Chard,~K.; Foster,~I.~T.; de~Pablo,~J.
  Blending education and polymer science: Semiautomated creation of a
  thermodynamic property database. \emph{Journal of chemical education}
  \textbf{2016}, \emph{93}, 1561--1568\relax
\mciteBstWouldAddEndPuncttrue
\mciteSetBstMidEndSepPunct{\mcitedefaultmidpunct}
{\mcitedefaultendpunct}{\mcitedefaultseppunct}\relax
\EndOfBibitem
\bibitem[Tchoua \latin{et~al.}(2019)Tchoua, Ajith, Hong, Ward, Chard, Belikov,
  Audus, Patel, de~Pablo, and Foster]{tchoua2019creating}
Tchoua,~R.~B.; Ajith,~A.; Hong,~Z.; Ward,~L.~T.; Chard,~K.; Belikov,~A.;
  Audus,~D.~J.; Patel,~S.; de~Pablo,~J.~J.; Foster,~I.~T. Creating training
  data for scientific named entity recognition with minimal human effort.
  International Conference on Computational Science. 2019; pp 398--411\relax
\mciteBstWouldAddEndPuncttrue
\mciteSetBstMidEndSepPunct{\mcitedefaultmidpunct}
{\mcitedefaultendpunct}{\mcitedefaultseppunct}\relax
\EndOfBibitem
\bibitem[Friedl(2006)]{friedl2006mastering}
Friedl,~J.~E. \emph{Mastering regular expressions}; " O'Reilly Media, Inc.",
  2006\relax
\mciteBstWouldAddEndPuncttrue
\mciteSetBstMidEndSepPunct{\mcitedefaultmidpunct}
{\mcitedefaultendpunct}{\mcitedefaultseppunct}\relax
\EndOfBibitem
\bibitem[Schwalbe-Koda \latin{et~al.}(2019)Schwalbe-Koda, Jensen, Olivetti, and
  G{\'o}mez-Bombarelli]{schwalbe2019graph}
Schwalbe-Koda,~D.; Jensen,~Z.; Olivetti,~E.; G{\'o}mez-Bombarelli,~R. Graph
  similarity drives zeolite diffusionless transformations and intergrowth.
  \emph{Nature materials} \textbf{2019}, \emph{18}, 1177--1181\relax
\mciteBstWouldAddEndPuncttrue
\mciteSetBstMidEndSepPunct{\mcitedefaultmidpunct}
{\mcitedefaultendpunct}{\mcitedefaultseppunct}\relax
\EndOfBibitem
\bibitem[Shetty and Ramprasad(2020)Shetty, and Ramprasad]{shettyautomated}
Shetty,~P.; Ramprasad,~R. Automated knowledge extraction from polymer
  literature using natural language processing. \emph{iScience} \textbf{2020},
  \emph{24}, 101922\relax
\mciteBstWouldAddEndPuncttrue
\mciteSetBstMidEndSepPunct{\mcitedefaultmidpunct}
{\mcitedefaultendpunct}{\mcitedefaultseppunct}\relax
\EndOfBibitem
\bibitem[Tshitoyan \latin{et~al.}(2019)Tshitoyan, Dagdelen, Weston, Dunn, Rong,
  Kononova, Persson, Ceder, and Jain]{tshitoyan2019unsupervised}
Tshitoyan,~V.; Dagdelen,~J.; Weston,~L.; Dunn,~A.; Rong,~Z.; Kononova,~O.;
  Persson,~K.~A.; Ceder,~G.; Jain,~A. Unsupervised word embeddings capture
  latent knowledge from materials science literature. \emph{Nature}
  \textbf{2019}, \emph{571}, 95--98\relax
\mciteBstWouldAddEndPuncttrue
\mciteSetBstMidEndSepPunct{\mcitedefaultmidpunct}
{\mcitedefaultendpunct}{\mcitedefaultseppunct}\relax
\EndOfBibitem
\bibitem[Gu \latin{et~al.}(2020)Gu, Tinn, Cheng, Lucas, Usuyama, Liu, Naumann,
  Gao, and Poon]{gu2020domain}
Gu,~Y.; Tinn,~R.; Cheng,~H.; Lucas,~M.; Usuyama,~N.; Liu,~X.; Naumann,~T.;
  Gao,~J.; Poon,~H. Domain-specific language model pretraining for biomedical
  natural language processing. \emph{arXiv preprint arXiv:2007.15779}
  \textbf{2020}, \relax
\mciteBstWouldAddEndPunctfalse
\mciteSetBstMidEndSepPunct{\mcitedefaultmidpunct}
{}{\mcitedefaultseppunct}\relax
\EndOfBibitem
\bibitem[Lee \latin{et~al.}(2020)Lee, Yoon, Kim, Kim, Kim, So, and
  Kang]{lee2020biobert}
Lee,~J.; Yoon,~W.; Kim,~S.; Kim,~D.; Kim,~S.; So,~C.~H.; Kang,~J. BioBERT: a
  pre-trained biomedical language representation model for biomedical text
  mining. \emph{Bioinformatics} \textbf{2020}, \emph{36}, 1234--1240\relax
\mciteBstWouldAddEndPuncttrue
\mciteSetBstMidEndSepPunct{\mcitedefaultmidpunct}
{\mcitedefaultendpunct}{\mcitedefaultseppunct}\relax
\EndOfBibitem
\bibitem[Guo \latin{et~al.}(2021)Guo, Ibanez-Lopez, Gao, Quach, Coley, Jensen,
  and Barzilay]{guo2021automated}
Guo,~J.; Ibanez-Lopez,~A.~S.; Gao,~H.; Quach,~V.; Coley,~C.~W.; Jensen,~K.~F.;
  Barzilay,~R. Automated Chemical Reaction Extraction from Scientific
  Literature. \emph{Journal of Chemical Information and Modeling}
  \textbf{2021}, \relax
\mciteBstWouldAddEndPunctfalse
\mciteSetBstMidEndSepPunct{\mcitedefaultmidpunct}
{}{\mcitedefaultseppunct}\relax
\EndOfBibitem
\bibitem[Weston \latin{et~al.}(2019)Weston, Tshitoyan, Dagdelen, Kononova,
  Trewartha, Persson, Ceder, and Jain]{weston2019named}
Weston,~L.; Tshitoyan,~V.; Dagdelen,~J.; Kononova,~O.; Trewartha,~A.;
  Persson,~K.~A.; Ceder,~G.; Jain,~A. Named entity recognition and
  normalization applied to large-scale information extraction from the
  materials science literature. \emph{Journal of chemical information and
  modeling} \textbf{2019}, \emph{59}, 3692--3702\relax
\mciteBstWouldAddEndPuncttrue
\mciteSetBstMidEndSepPunct{\mcitedefaultmidpunct}
{\mcitedefaultendpunct}{\mcitedefaultseppunct}\relax
\EndOfBibitem
\bibitem[Fleiss(1971)]{fleiss1971measuring}
Fleiss,~J.~L. Measuring nominal scale agreement among many raters.
  \emph{Psychological bulletin} \textbf{1971}, \emph{76}, 378\relax
\mciteBstWouldAddEndPuncttrue
\mciteSetBstMidEndSepPunct{\mcitedefaultmidpunct}
{\mcitedefaultendpunct}{\mcitedefaultseppunct}\relax
\EndOfBibitem
\bibitem[Tabassum \latin{et~al.}(2020)Tabassum, Lee, Xu, and
  Ritter]{tabassum2020wnut}
Tabassum,~J.; Lee,~S.; Xu,~W.; Ritter,~A. WNUT-2020 task 1 overview: Extracting
  entities and relations from wet lab protocols. \emph{arXiv preprint
  arXiv:2010.14576} \textbf{2020}, \relax
\mciteBstWouldAddEndPunctfalse
\mciteSetBstMidEndSepPunct{\mcitedefaultmidpunct}
{}{\mcitedefaultseppunct}\relax
\EndOfBibitem
\bibitem[Liang \latin{et~al.}(2020)Liang, Yu, Jiang, Er, Wang, Zhao, and
  Zhang]{liang2020bond}
Liang,~C.; Yu,~Y.; Jiang,~H.; Er,~S.; Wang,~R.; Zhao,~T.; Zhang,~C. Bond:
  Bert-assisted open-domain named entity recognition with distant supervision.
  Proceedings of the 26th ACM SIGKDD International Conference on Knowledge
  Discovery \& Data Mining. 2020; pp 1054--1064\relax
\mciteBstWouldAddEndPuncttrue
\mciteSetBstMidEndSepPunct{\mcitedefaultmidpunct}
{\mcitedefaultendpunct}{\mcitedefaultseppunct}\relax
\EndOfBibitem
\bibitem[Trewartha \latin{et~al.}(2022)Trewartha, Walker, Huo, Lee, Cruse,
  Dagdelen, Dunn, Persson, Ceder, and Jain]{trewartha2022quantifying}
Trewartha,~A.; Walker,~N.; Huo,~H.; Lee,~S.; Cruse,~K.; Dagdelen,~J.; Dunn,~A.;
  Persson,~K.~A.; Ceder,~G.; Jain,~A. Quantifying the advantage of
  domain-specific pre-training on named entity recognition tasks in materials
  science. \emph{Patterns} \textbf{2022}, \emph{3}, 100488\relax
\mciteBstWouldAddEndPuncttrue
\mciteSetBstMidEndSepPunct{\mcitedefaultmidpunct}
{\mcitedefaultendpunct}{\mcitedefaultseppunct}\relax
\EndOfBibitem
\bibitem[Huang \latin{et~al.}(2015)Huang, Xu, and Yu]{huang2015bidirectional}
Huang,~Z.; Xu,~W.; Yu,~K. Bidirectional LSTM-CRF models for sequence tagging.
  \emph{arXiv preprint arXiv:1508.01991} \textbf{2015}, \relax
\mciteBstWouldAddEndPunctfalse
\mciteSetBstMidEndSepPunct{\mcitedefaultmidpunct}
{}{\mcitedefaultseppunct}\relax
\EndOfBibitem
\bibitem[Krallinger \latin{et~al.}(2015)Krallinger, Rabal, Leitner, Vazquez,
  Salgado, Lu, Leaman, Lu, Ji, Lowe, \latin{et~al.}
  others]{krallinger2015chemdner}
Krallinger,~M.; Rabal,~O.; Leitner,~F.; Vazquez,~M.; Salgado,~D.; Lu,~Z.;
  Leaman,~R.; Lu,~Y.; Ji,~D.; Lowe,~D.~M., \latin{et~al.}  The CHEMDNER corpus
  of chemicals and drugs and its annotation principles. \emph{Journal of
  cheminformatics} \textbf{2015}, \emph{7}, 1--17\relax
\mciteBstWouldAddEndPuncttrue
\mciteSetBstMidEndSepPunct{\mcitedefaultmidpunct}
{\mcitedefaultendpunct}{\mcitedefaultseppunct}\relax
\EndOfBibitem
\bibitem[Mysore \latin{et~al.}(2019)Mysore, Jensen, Kim, Huang, Chang,
  Strubell, Flanigan, McCallum, and Olivetti]{mysore2019materials}
Mysore,~S.; Jensen,~Z.; Kim,~E.; Huang,~K.; Chang,~H.-S.; Strubell,~E.;
  Flanigan,~J.; McCallum,~A.; Olivetti,~E. The materials science procedural
  text corpus: Annotating materials synthesis procedures with shallow semantic
  structures. \emph{arXiv preprint arXiv:1905.06939} \textbf{2019}, \relax
\mciteBstWouldAddEndPunctfalse
\mciteSetBstMidEndSepPunct{\mcitedefaultmidpunct}
{}{\mcitedefaultseppunct}\relax
\EndOfBibitem
\bibitem[Otsuka \latin{et~al.}(2011)Otsuka, Kuwajima, Hosoya, Xu, and
  Yamazaki]{otsuka2011polyinfo}
Otsuka,~S.; Kuwajima,~I.; Hosoya,~J.; Xu,~Y.; Yamazaki,~M. PoLyInfo: Polymer
  database for polymeric materials design. 2011 International Conference on
  Emerging Intelligent Data and Web Technologies. 2011; pp 22--29\relax
\mciteBstWouldAddEndPuncttrue
\mciteSetBstMidEndSepPunct{\mcitedefaultmidpunct}
{\mcitedefaultendpunct}{\mcitedefaultseppunct}\relax
\EndOfBibitem
\bibitem[Shetty and Ramprasad(2021)Shetty, and Ramprasad]{shetty2021machine}
Shetty,~P.; Ramprasad,~R. Machine-Guided Polymer Knowledge Extraction Using
  Natural Language Processing: The Example of Named Entity Normalization.
  \emph{Journal of Chemical Information and Modeling} \textbf{2021}, \emph{61},
  5377--5385\relax
\mciteBstWouldAddEndPuncttrue
\mciteSetBstMidEndSepPunct{\mcitedefaultmidpunct}
{\mcitedefaultendpunct}{\mcitedefaultseppunct}\relax
\EndOfBibitem
\bibitem[Palomba \latin{et~al.}(2014)Palomba, Vazquez, and
  D{\'\i}az]{palomba2014prediction}
Palomba,~D.; Vazquez,~G.~E.; D{\'\i}az,~M.~F. Prediction of elongation at break
  for linear polymers. \emph{Chemometrics and Intelligent Laboratory Systems}
  \textbf{2014}, \emph{139}, 121--131\relax
\mciteBstWouldAddEndPuncttrue
\mciteSetBstMidEndSepPunct{\mcitedefaultmidpunct}
{\mcitedefaultendpunct}{\mcitedefaultseppunct}\relax
\EndOfBibitem
\bibitem[Doan~Tran \latin{et~al.}(2020)Doan~Tran, Kim, Chen, Chandrasekaran,
  Batra, Venkatram, Kamal, Lightstone, Gurnani, Shetty, \latin{et~al.}
  others]{pg}
Doan~Tran,~H.; Kim,~C.; Chen,~L.; Chandrasekaran,~A.; Batra,~R.; Venkatram,~S.;
  Kamal,~D.; Lightstone,~J.~P.; Gurnani,~R.; Shetty,~P., \latin{et~al.}
  Machine-learning predictions of polymer properties with Polymer Genome.
  \emph{Journal of Applied Physics} \textbf{2020}, \emph{128}, 171104\relax
\mciteBstWouldAddEndPuncttrue
\mciteSetBstMidEndSepPunct{\mcitedefaultmidpunct}
{\mcitedefaultendpunct}{\mcitedefaultseppunct}\relax
\EndOfBibitem
\bibitem[Zhang \latin{et~al.}(2018)Zhang, Niu, and Wu]{zhang2018polyimide}
Zhang,~M.; Niu,~H.; Wu,~D. Polyimide fibers with high strength and high
  modulus: preparation, structures, properties, and applications.
  \emph{Macromolecular rapid communications} \textbf{2018}, \emph{39},
  1800141\relax
\mciteBstWouldAddEndPuncttrue
\mciteSetBstMidEndSepPunct{\mcitedefaultmidpunct}
{\mcitedefaultendpunct}{\mcitedefaultseppunct}\relax
\EndOfBibitem
\bibitem[Wang \latin{et~al.}(2020)Wang, Zhang, Zhang, Xu, and
  Zhang]{wang2020evading}
Wang,~C.; Zhang,~S.; Zhang,~L.; Xu,~Y.; Zhang,~L. Evading the
  strength--ductility trade-off dilemma of rigid thermosets by incorporating
  triple cross-links of varying strengths. \emph{Polymer Chemistry}
  \textbf{2020}, \emph{11}, 6281--6287\relax
\mciteBstWouldAddEndPuncttrue
\mciteSetBstMidEndSepPunct{\mcitedefaultmidpunct}
{\mcitedefaultendpunct}{\mcitedefaultseppunct}\relax
\EndOfBibitem
\bibitem[Nagasawa \latin{et~al.}(2018)Nagasawa, Al-Naamani, and
  Saeki]{nagasawa2018computer}
Nagasawa,~S.; Al-Naamani,~E.; Saeki,~A. Computer-aided screening of conjugated
  polymers for organic solar cell: classification by random forest. \emph{The
  Journal of Physical Chemistry Letters} \textbf{2018}, \emph{9},
  2639--2646\relax
\mciteBstWouldAddEndPuncttrue
\mciteSetBstMidEndSepPunct{\mcitedefaultmidpunct}
{\mcitedefaultendpunct}{\mcitedefaultseppunct}\relax
\EndOfBibitem
\bibitem[Zhang and Li(2021)Zhang, and Li]{psc}
Zhang,~Z.-G.; Li,~Y. Polymerized small-molecule acceptors for high-performance
  all-polymer solar cells. \emph{Angewandte Chemie International Edition}
  \textbf{2021}, \emph{60}, 4422--4433\relax
\mciteBstWouldAddEndPuncttrue
\mciteSetBstMidEndSepPunct{\mcitedefaultmidpunct}
{\mcitedefaultendpunct}{\mcitedefaultseppunct}\relax
\EndOfBibitem
\bibitem[Zhang \latin{et~al.}(2022)Zhang, Ma, Zhang, Zhu, Xu, Zhang, Tsang,
  Lee, Woo, He, \latin{et~al.} others]{zhang2022metallated}
Zhang,~M.; Ma,~X.; Zhang,~H.; Zhu,~L.; Xu,~L.; Zhang,~F.; Tsang,~C.-S.; Lee,~L.
  Y.~S.; Woo,~H.~Y.; He,~Z., \latin{et~al.}  Metallated terpolymer donors with
  strongly absorbing iridium complex enables polymer solar cells with 16.71\%
  efficiency. \emph{Chemical Engineering Journal} \textbf{2022}, \emph{430},
  132832\relax
\mciteBstWouldAddEndPuncttrue
\mciteSetBstMidEndSepPunct{\mcitedefaultmidpunct}
{\mcitedefaultendpunct}{\mcitedefaultseppunct}\relax
\EndOfBibitem
\bibitem[Abdelkareem \latin{et~al.}(2021)Abdelkareem, Elsaid, Wilberforce,
  Kamil, Sayed, and Olabi]{abdelkareem2021environmental}
Abdelkareem,~M.~A.; Elsaid,~K.; Wilberforce,~T.; Kamil,~M.; Sayed,~E.~T.;
  Olabi,~A. Environmental aspects of fuel cells: A review. \emph{Science of The
  Total Environment} \textbf{2021}, \emph{752}, 141803\relax
\mciteBstWouldAddEndPuncttrue
\mciteSetBstMidEndSepPunct{\mcitedefaultmidpunct}
{\mcitedefaultendpunct}{\mcitedefaultseppunct}\relax
\EndOfBibitem
\bibitem[Larminie \latin{et~al.}(2003)Larminie, Dicks, and
  McDonald]{larminie2003fuel}
Larminie,~J.; Dicks,~A.; McDonald,~M.~S. \emph{Fuel cell systems explained}; J.
  Wiley Chichester, UK, 2003; Vol.~2\relax
\mciteBstWouldAddEndPuncttrue
\mciteSetBstMidEndSepPunct{\mcitedefaultmidpunct}
{\mcitedefaultendpunct}{\mcitedefaultseppunct}\relax
\EndOfBibitem
\bibitem[Shaari \latin{et~al.}(2018)Shaari, Kamarudin, Basri, Shyuan, Masdar,
  and Nordin]{shaari2018enhancedMP}
Shaari,~N.; Kamarudin,~S.; Basri,~S.; Shyuan,~L.; Masdar,~M.; Nordin,~D.
  Enhanced proton conductivity and methanol permeability reduction via sodium
  alginate electrolyte-sulfonated graphene oxide bio-membrane. \emph{Nanoscale
  research letters} \textbf{2018}, \emph{13}, 1--16\relax
\mciteBstWouldAddEndPuncttrue
\mciteSetBstMidEndSepPunct{\mcitedefaultmidpunct}
{\mcitedefaultendpunct}{\mcitedefaultseppunct}\relax
\EndOfBibitem
\bibitem[Catenaro \latin{et~al.}(2021)Catenaro, Rizzo, and
  Onori]{catenaro2021ragone}
Catenaro,~E.; Rizzo,~D.~M.; Onori,~S. Experimental analysis and analytical
  modeling of enhanced-Ragone plot. \emph{Applied Energy} \textbf{2021},
  \emph{291}, 116473\relax
\mciteBstWouldAddEndPuncttrue
\mciteSetBstMidEndSepPunct{\mcitedefaultmidpunct}
{\mcitedefaultendpunct}{\mcitedefaultseppunct}\relax
\EndOfBibitem
\bibitem[Shown \latin{et~al.}(2015)Shown, Ganguly, Chen, and
  Chen]{shown2015conducting}
Shown,~I.; Ganguly,~A.; Chen,~L.-C.; Chen,~K.-H. Conducting polymer-based
  flexible supercapacitor. \emph{Energy Science \& Engineering} \textbf{2015},
  \emph{3}, 2--26\relax
\mciteBstWouldAddEndPuncttrue
\mciteSetBstMidEndSepPunct{\mcitedefaultmidpunct}
{\mcitedefaultendpunct}{\mcitedefaultseppunct}\relax
\EndOfBibitem
\bibitem[Uppugalla and Srinivasan(2019)Uppugalla, and
  Srinivasan]{uppugalla2019polyaniline}
Uppugalla,~S.; Srinivasan,~P. Polyaniline nanofibers and porous Ni [OH] 2
  sheets coated carbon fabric for high performance super capacitor.
  \emph{Journal of Applied Polymer Science} \textbf{2019}, \emph{136},
  48042\relax
\mciteBstWouldAddEndPuncttrue
\mciteSetBstMidEndSepPunct{\mcitedefaultmidpunct}
{\mcitedefaultendpunct}{\mcitedefaultseppunct}\relax
\EndOfBibitem
\bibitem[Li \latin{et~al.}(2019)Li, Yang, Zhou, Lin, Xu, Xing, Zhang, Feng, Mu,
  Li, \latin{et~al.} others]{li2019coupled}
Li,~Y.; Yang,~Y.; Zhou,~J.; Lin,~S.; Xu,~Z.; Xing,~Y.; Zhang,~Y.; Feng,~J.;
  Mu,~Z.; Li,~P., \latin{et~al.}  Coupled and decoupled hierarchical carbon
  nanomaterials toward high-energy-density quasi-solid-state Na-Ion hybrid
  energy storage devices. \emph{Energy Storage Materials} \textbf{2019},
  \emph{23}, 530--538\relax
\mciteBstWouldAddEndPuncttrue
\mciteSetBstMidEndSepPunct{\mcitedefaultmidpunct}
{\mcitedefaultendpunct}{\mcitedefaultseppunct}\relax
\EndOfBibitem
\bibitem[Javed \latin{et~al.}(2019)Javed, Lei, Shah, Asim, Raza, and
  Mai]{javed2019achieving}
Javed,~M.~S.; Lei,~H.; Shah,~H.~U.; Asim,~S.; Raza,~R.; Mai,~W. Achieving high
  rate and high energy density in an all-solid-state flexible asymmetric
  pseudocapacitor through the synergistic design of binder-free 3D ZnCo 2 O 4
  nano polyhedra and 2D layered Ti 3 C 2 T x-MXenes. \emph{Journal of Materials
  Chemistry A} \textbf{2019}, \emph{7}, 24543--24556\relax
\mciteBstWouldAddEndPuncttrue
\mciteSetBstMidEndSepPunct{\mcitedefaultmidpunct}
{\mcitedefaultendpunct}{\mcitedefaultseppunct}\relax
\EndOfBibitem
\bibitem[Fu \latin{et~al.}(2019)Fu, Wang, and Sun]{fu2019polymer}
Fu,~H.; Wang,~Z.; Sun,~Y. Polymer donors for high-performance non-fullerene
  organic solar cells. \emph{Angewandte Chemie International Edition}
  \textbf{2019}, \emph{58}, 4442--4453\relax
\mciteBstWouldAddEndPuncttrue
\mciteSetBstMidEndSepPunct{\mcitedefaultmidpunct}
{\mcitedefaultendpunct}{\mcitedefaultseppunct}\relax
\EndOfBibitem
\bibitem[Ciotti \latin{et~al.}(2020)Ciotti, Ciccozzi, Terrinoni, Jiang, Wang,
  and Bernardini]{ciotti2020covid}
Ciotti,~M.; Ciccozzi,~M.; Terrinoni,~A.; Jiang,~W.-C.; Wang,~C.-B.;
  Bernardini,~S. The COVID-19 pandemic. \emph{Critical reviews in clinical
  laboratory sciences} \textbf{2020}, \emph{57}, 365--388\relax
\mciteBstWouldAddEndPuncttrue
\mciteSetBstMidEndSepPunct{\mcitedefaultmidpunct}
{\mcitedefaultendpunct}{\mcitedefaultseppunct}\relax
\EndOfBibitem
\bibitem[Gao \latin{et~al.}(2021)Gao, Yin, Myers, Lakhani, and
  Wang]{gao2021potentially}
Gao,~J.; Yin,~Y.; Myers,~K.~R.; Lakhani,~K.~R.; Wang,~D. Potentially
  long-lasting effects of the pandemic on scientists. \emph{Nature
  communications} \textbf{2021}, \emph{12}, 1--6\relax
\mciteBstWouldAddEndPuncttrue
\mciteSetBstMidEndSepPunct{\mcitedefaultmidpunct}
{\mcitedefaultendpunct}{\mcitedefaultseppunct}\relax
\EndOfBibitem
\bibitem[Weininger \latin{et~al.}(1989)Weininger, Weininger, and
  Weininger]{weininger1989smiles}
Weininger,~D.; Weininger,~A.; Weininger,~J.~L. SMILES. 2. Algorithm for
  generation of unique SMILES notation. \emph{Journal of chemical information
  and computer sciences} \textbf{1989}, \emph{29}, 97--101\relax
\mciteBstWouldAddEndPuncttrue
\mciteSetBstMidEndSepPunct{\mcitedefaultmidpunct}
{\mcitedefaultendpunct}{\mcitedefaultseppunct}\relax
\EndOfBibitem
\bibitem[Wu \latin{et~al.}(2016)Wu, Schuster, Chen, Le, Norouzi, Macherey,
  Krikun, Cao, Gao, Macherey, \latin{et~al.} others]{wu2016google}
Wu,~Y.; Schuster,~M.; Chen,~Z.; Le,~Q.~V.; Norouzi,~M.; Macherey,~W.;
  Krikun,~M.; Cao,~Y.; Gao,~Q.; Macherey,~K., \latin{et~al.}  Google's neural
  machine translation system: Bridging the gap between human and machine
  translation. \emph{arXiv preprint arXiv:1609.08144} \textbf{2016}, \relax
\mciteBstWouldAddEndPunctfalse
\mciteSetBstMidEndSepPunct{\mcitedefaultmidpunct}
{}{\mcitedefaultseppunct}\relax
\EndOfBibitem
\bibitem[Song \latin{et~al.}(2020)Song, Salcianu, Song, Dopson, and
  Zhou]{song2020linear}
Song,~X.; Salcianu,~A.; Song,~Y.; Dopson,~D.; Zhou,~D. Linear-Time WordPiece
  Tokenization. \emph{arXiv e-prints} \textbf{2020}, arXiv--2012\relax
\mciteBstWouldAddEndPuncttrue
\mciteSetBstMidEndSepPunct{\mcitedefaultmidpunct}
{\mcitedefaultendpunct}{\mcitedefaultseppunct}\relax
\EndOfBibitem
\bibitem[Zhu \latin{et~al.}(2015)Zhu, Kiros, Zemel, Salakhutdinov, Urtasun,
  Torralba, and Fidler]{zhu2015aligning}
Zhu,~Y.; Kiros,~R.; Zemel,~R.; Salakhutdinov,~R.; Urtasun,~R.; Torralba,~A.;
  Fidler,~S. Aligning books and movies: Towards story-like visual explanations
  by watching movies and reading books. Proceedings of the IEEE international
  conference on computer vision. 2015; pp 19--27\relax
\mciteBstWouldAddEndPuncttrue
\mciteSetBstMidEndSepPunct{\mcitedefaultmidpunct}
{\mcitedefaultendpunct}{\mcitedefaultseppunct}\relax
\EndOfBibitem
\bibitem[Liu \latin{et~al.}(2019)Liu, Ott, Goyal, Du, Joshi, Chen, Levy, Lewis,
  Zettlemoyer, and Stoyanov]{liu2019roberta}
Liu,~Y.; Ott,~M.; Goyal,~N.; Du,~J.; Joshi,~M.; Chen,~D.; Levy,~O.; Lewis,~M.;
  Zettlemoyer,~L.; Stoyanov,~V. Roberta: A robustly optimized bert pretraining
  approach. \emph{arXiv preprint arXiv:1907.11692} \textbf{2019}, \relax
\mciteBstWouldAddEndPunctfalse
\mciteSetBstMidEndSepPunct{\mcitedefaultmidpunct}
{}{\mcitedefaultseppunct}\relax
\EndOfBibitem
\bibitem[Wolf \latin{et~al.}(2020)Wolf, Debut, Sanh, Chaumond, Delangue, Moi,
  Cistac, Rault, Louf, Funtowicz, Davison, Shleifer, von Platen, Ma, Jernite,
  Plu, Xu, Le~Scao, Gugger, Drame, Lhoest, and
  Rush]{wolf-etal-2020-transformers}
Wolf,~T. \latin{et~al.}  Transformers: State-of-the-Art Natural Language
  Processing. Proceedings of the 2020 Conference on Empirical Methods in
  Natural Language Processing: System Demonstrations. Online, 2020; pp
  38--45\relax
\mciteBstWouldAddEndPuncttrue
\mciteSetBstMidEndSepPunct{\mcitedefaultmidpunct}
{\mcitedefaultendpunct}{\mcitedefaultseppunct}\relax
\EndOfBibitem
\bibitem[Huang \latin{et~al.}(2019)Huang, Altosaar, and
  Ranganath]{huang2019clinicalbert}
Huang,~K.; Altosaar,~J.; Ranganath,~R. Clinicalbert: Modeling clinical notes
  and predicting hospital readmission. \emph{arXiv preprint arXiv:1904.05342}
  \textbf{2019}, \relax
\mciteBstWouldAddEndPunctfalse
\mciteSetBstMidEndSepPunct{\mcitedefaultmidpunct}
{}{\mcitedefaultseppunct}\relax
\EndOfBibitem
\bibitem[Araci(2019)]{araci2019finbert}
Araci,~D. Finbert: Financial sentiment analysis with pre-trained language
  models. \emph{arXiv preprint arXiv:1908.10063} \textbf{2019}, \relax
\mciteBstWouldAddEndPunctfalse
\mciteSetBstMidEndSepPunct{\mcitedefaultmidpunct}
{}{\mcitedefaultseppunct}\relax
\EndOfBibitem
\bibitem[Levenshtein(1966)]{levenshtein}
Levenshtein,~V.~I. Binary codes capable of correcting deletions, insertions,
  and reversals. Soviet physics doklady. 1966; pp 707--710\relax
\mciteBstWouldAddEndPuncttrue
\mciteSetBstMidEndSepPunct{\mcitedefaultmidpunct}
{\mcitedefaultendpunct}{\mcitedefaultseppunct}\relax
\EndOfBibitem
\bibitem[Mitkov(2014)]{mitkov2014anaphora}
Mitkov,~R. \emph{Anaphora resolution}; Routledge, 2014\relax
\mciteBstWouldAddEndPuncttrue
\mciteSetBstMidEndSepPunct{\mcitedefaultmidpunct}
{\mcitedefaultendpunct}{\mcitedefaultseppunct}\relax
\EndOfBibitem
\bibitem[Wang \latin{et~al.}(2016)Wang, Cao, De~Melo, and
  Liu]{wang2016relation}
Wang,~L.; Cao,~Z.; De~Melo,~G.; Liu,~Z. Relation classification via multi-level
  attention cnns. Proceedings of the 54th Annual Meeting of the Association for
  Computational Linguistics (Volume 1: Long Papers). 2016; pp 1298--1307\relax
\mciteBstWouldAddEndPuncttrue
\mciteSetBstMidEndSepPunct{\mcitedefaultmidpunct}
{\mcitedefaultendpunct}{\mcitedefaultseppunct}\relax
\EndOfBibitem
\bibitem[Zhou \latin{et~al.}(2020)Zhou, Lin, Lin, Wang, Du, Neves, and
  Ren]{zhou2020nero}
Zhou,~W.; Lin,~H.; Lin,~B.~Y.; Wang,~Z.; Du,~J.; Neves,~L.; Ren,~X. Nero: A
  neural rule grounding framework for label-efficient relation extraction.
  Proceedings of The Web Conference 2020. 2020; pp 2166--2176\relax
\mciteBstWouldAddEndPuncttrue
\mciteSetBstMidEndSepPunct{\mcitedefaultmidpunct}
{\mcitedefaultendpunct}{\mcitedefaultseppunct}\relax
\EndOfBibitem
\bibitem[Williams and Rasmussen(2006)Williams, and
  Rasmussen]{williams2006gaussian}
Williams,~C.~K.; Rasmussen,~C.~E. \emph{Gaussian processes for machine
  learning}; MIT press Cambridge, MA, 2006; Vol.~2\relax
\mciteBstWouldAddEndPuncttrue
\mciteSetBstMidEndSepPunct{\mcitedefaultmidpunct}
{\mcitedefaultendpunct}{\mcitedefaultseppunct}\relax
\EndOfBibitem
\bibitem[Chai and Draxler(2014)Chai, and Draxler]{chai2014root}
Chai,~T.; Draxler,~R.~R. Root mean square error (RMSE) or mean absolute error
  (MAE). \emph{Geoscientific Model Development Discussions} \textbf{2014},
  \emph{7}, 1525--1534\relax
\mciteBstWouldAddEndPuncttrue
\mciteSetBstMidEndSepPunct{\mcitedefaultmidpunct}
{\mcitedefaultendpunct}{\mcitedefaultseppunct}\relax
\EndOfBibitem
\end{mcitethebibliography}

\newpage

\setcounter{figure}{0}
\renewcommand{\thefigure}{S\arabic{figure}}
\setcounter{table}{0}
\renewcommand{\thetable}{S\arabic{table}}

\begin{center}
\Huge
\title{Supplementary Information}
\end{center}

\section{S1. Annotation guidelines}

We annotated abstracts that had material property information and hence the entity types defined in our ontology. We only labeled material entities that were explicitly part of the material formulation for which a property value pair was reported in the abstract. This was done in order to encode some notion of jointly extracting entities and relationships in the model through minimal annotation effort.

\begin{enumerate}
    \item POLYMER: All polymer material entities that occur in a material formulation for which property values are reported in the abstract should be labeled. Abbreviations when found next to it in brackets should also be labeled (excluding brackets). This includes homopolymers, copolymers and blends.

    This should refer to a polymer name rather than a generic polymer family. If annotating multiple polymer entries separated by a forward slash e.g. polyethylene/polypropylene, tag each material entity on either side of the slash separately. End-functionalization of a polymer can be omitted from the labeled token.

    \item POLYMER\_FAMILY: All polymeric material entities that refer only to polymer families (e.g.: polyamide, polyimide, polybenzimadozole etc) that occur in a material formulation for which property values are reported in the abstract should be labeled. Abbreviations when found next to it in brackets should also be labeled (excluding brackets).
 
    \item ORGANIC: Label all organic compounds that are not polymers and are used in the material formulation for which property values are reported. These could be organic molecules used as plasticizers, crosslinkers, blended with the polymer or as grafts. These are explicitly a part of the material formulation and the extracted record should include this information.
 
    \item MONOMER: Label all monomer repeat units for polymers for which property values are reported in the abstract. These are being annotated separately as they are not explicitly part of the material formulation of interest but are nevertheless commonly reported in abstracts. It is hence necessary to distinguish between ORGANIC additives and monomers. Look for contextual cues such as `synthesized from', `prepared from' or the presence of the word monomer.

    \item INORGANIC: Label all inorganic materials (such as SiO\textsubscript{2}, TiO\textsubscript{2} etc) explicitly used in a material formulation.

    \item MATERIAL\_AMT: Numerical quantity + unit denoting the amount of a material like additive or blend for which a property value is reported. This is typically wt \% or mol.

    \item PROPERTY\_NAME: Name of property measured. Label properties for which a corresponding numeric value is reported in the abstract. Label the property name and abbreviation in bracket (if present) separately.

    \item PROPERTY\_VALUE: Numeric value + unit for a reported PROPERTY\_NAME.
\end{enumerate}

\section{S2. NER Datasets used for testing BERT-based encoders}

\begin{enumerate}
    \item \textbf{ChemDNER}\cite{krallinger2015chemdner}: This is a dataset of 10,000 PubMed abstracts annotated for chemical entity mentions using 7 different entity types such as ABBREVIATION, SYSTEMATIC, FORMULA, TRIVIAL, FAMILY, MULTIPLE, and IDENTIFIERS. The dataset is split into 3500 abstracts for train, 3500 validation abstracts, and 3000 test abstracts.
    \item \textbf{Inorganic Synthesis recipes}\cite{mysore2019materials}: This is a dataset of 230 inorganic synthesis paragraphs annotated using 21 entity types that are relevant in the context of materials synthesis. 15 paragraphs each are used for validation and testing and the remaining 200 paragraphs are used for training. 
    \item \textbf{Inorganic Abstracts}\cite{weston2019named}: This is a dataset of 800 abstracts related to inorganic materials in which 8 different entity types related to inorganic materials are annotated namely, inorganic material (MAT), symmetry/phase label (SPL), sample descriptor (DSC), material property (PRO), material application (APL), synthesis method (SMT), and characterization method (CMT). The dataset is split as 640/80/80 for training, validation, and testing respectively.
    \item \textbf{ChemRxnExtractor}\cite{guo2021automated}: This is a dataset of 329 organic synthesis paragraphs in which the product of the synthesis is labeled using the BIO scheme. The dataset is split into 251, 41, and 37 paragraphs in the train, validation, and test set respectively.
\end{enumerate}

\section{S3. Performance of MaterialsBERT on annotated polymer abstracts NER dataset}

\begin{table}[!h]
\caption{Performance of an NER model using MaterialsBERT as the encoder across various entity types in the ontology used in this work on the test set of polymer abstracts. Values are reported in \%}
\begin{center}
\begin{tabular}{|c|c|c|c|c|}
    \hline
    \textbf{Entity type} & \textbf{Precision} & \textbf{Recall} & \textbf{F1} & \textbf{Occurrence}\\
    \hline
    POLYMER & 75.9 & 83.8 & 79.6 & 8814\\
    \hline
    PROPERTY\_VALUE & 73.0 & 80.0 & 76.4 & 5749\\
    \hline
    PROPERTY\_NAME & 72.6 & 74.7 & 73.6 & 4486\\
    \hline
    ORGANIC & 34.7 & 21.0 & 26.2 & 2173\\
    \hline
    MONOMER & 67.1 & 77.6 & 72.0 & 2073\\
    \hline
    POLYMER\_FAMILY & 41.6 & 52.1 & 46.2 & 1476 \\
    \hline
    INORGANIC & 40.2 & 64.7 & 49.6 & 1261\\
    \hline
    MATERIAL\_AMOUNT & 59.1 & 83.9 & 79.6 & 1140\\
    \hline
\end{tabular}
\end{center}
\label{tab:entity_metrics}
\end{table}

The performance of the various entity types for MaterialsBERT, the best-performing model along with the distribution of various entity types in the annotated dataset of polymer abstracts is shown in Table \ref{tab:entity_metrics}. It is clear that there is a positive correlation between the number of occurrences of each entity type and the performance of the corresponding entity type. The ORGANIC entity type has a low F1 score likely because it is difficult to distinguish between MONOMER and ORGANIC entity types. The POLYMER\_FAMILY entity type has similarities with the POLYMER entity type which likely lowers the F1 score for this entity type as well.

\section{S4. Details of extracted data}

The number of entities extracted (accounting for variations in case) from the $\sim 300,000$ material property records is shown in Table \ref{tab:entities_extracted}. Note that these material entities are not unique materials as for instance, `silica' and `SiO\textsubscript{2}' would be counted as separate entities but are the same material. Similarly for property names, `\Tg' and `glass transition temperature' are counted as separate entities, even though they correspond to the same property.

\begin{table}[!h]
    \centering
    \begin{tabular}{|c|c|}
    \hline
         Entity type & Number of extracted entities \\
         \hline
         POLYMER & 74396 \\
         \hline
         MONOMER & 50000 \\
         \hline
         INORGANIC & 26114 \\
         \hline
         POLYMER\_FAMILY & 5109 \\
         \hline
         ORGANIC & 2769 \\
         \hline
         PROPERTY\_NAME & 33642 \\
         \hline
         
    \end{tabular}
    \caption{Number of entities extracted from the corpus of polymer relevant abstracts for key entity types from the ontology used in this work}
    \label{tab:entities_extracted}
\end{table}

\begin{figure}[!h]
    \centering
    \includegraphics[scale=0.13]{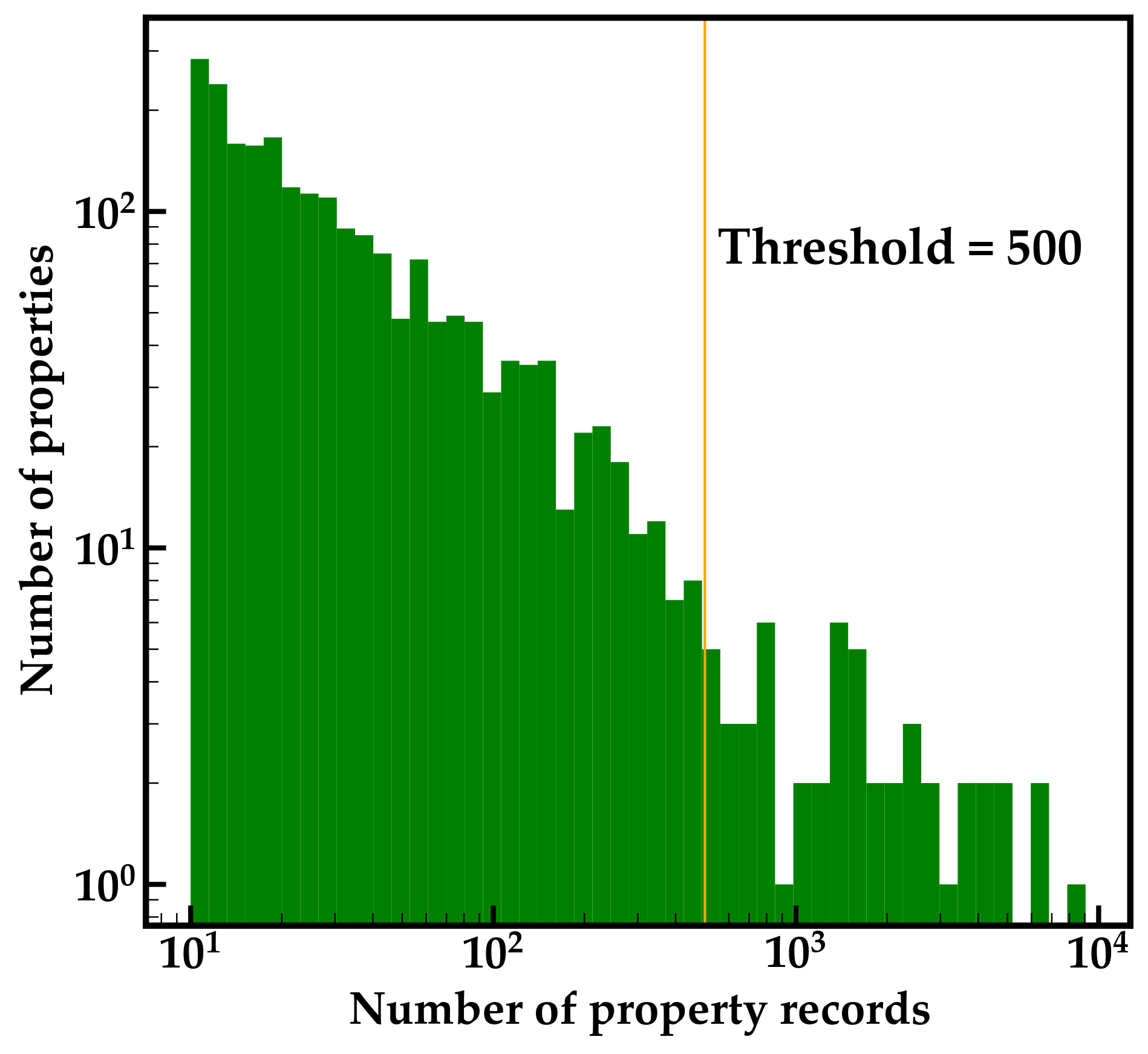}
    \caption{Histogram of number of material property records for each property entity. The threshold represents the cut-off above which the number of material property records is reported for each property in Table \ref{tab:prop_records_elaborate}}
    \label{fig:property_frequency_histogram}
\end{figure}

The histogram of number of records extracted for each property is shown in Figure \ref{fig:property_frequency_histogram}. Only property entities with at least 10 material property records are shown in this plot. Observe that the number of material property records appears to be power law distributed with a small number of properties having a large number of records associated with them while the majority of property entities only have a small number of associated records. We threshold this plot at 500 material property records and report the number of records associated with all properties above this threshold in Table \ref{tab:prop_records_elaborate}. Note that common variations in names for each property such as abbreviations (e.g. IEC for ion exchange capacity) or plural forms (e.g. activation energies) were accounted for in order to compute the number of records. These account for a total of 111279 records, i.e., 37 \% of $\sim 300,000$ material property records. Note that device properties such as power conversion efficiency, open circuit voltage etc listed in Table 5 of the main paper are listed in Table \ref{tab:prop_records_elaborate} as well but are more in number as they are not restricted to any particular application.

\begin{longtable}{|c|c|c|}
    \hline
         Index & Property name & Number of material property records \\
         \hline
               1 & Molecular weight & 9053      \\
         \hline
               2 & Power conversion efficiency & 8096    \\
        \hline
               3 & Glass transition temperature & 6155    \\

       \hline
               4 & Electrical conductivity & 6030      \\
        \hline
               5 & Ionic conductivity & 4933      \\
         \hline
               6 & Specific capacitance & 4916      \\
        \hline
               7 & Tensile strength & 4382      \\
        \hline
               8 & Number average molecular weight & 4096      \\
         \hline
            9 &    Water contact angle & 3932      \\
         \hline
              10 &  Discharge capacity & 3632      \\
         \hline
               11 & Polydispersity index & 3087      \\
         \hline
               12 & Specific surface area & 2912    \\
         \hline
               13 & Gravimetric power density & 2816      \\
        \hline
              14 &  Adsorption capacity & 2555      \\
         \hline
               15 & Activation energy & 2408      \\
        \hline
              16 &  Open circuit voltage & 2343      \\
         \hline
         17 &    Band gap & 2245      \\
         \hline
         18 &    Viscosity & 2084      \\
         \hline
            19 & Youngs modulus & 1904      \\
         \hline
              20 &  Gravimetric energy density & 1765      \\
        \hline
              21 &  Short circuit current & 1699      \\
        \hline
              22 &  Melting temperature & 1615      \\
         \hline
            23 &    Sensitivity & 1602      \\
         \hline
              24 &  Dielectric constant & 1534      \\
          \hline
              25 &  Elongation at break & 1499      \\
        \hline
              26 &  Thermal Decomposition Temperature & 1479      \\
        
        \hline
              27 &  Fill factor & 1431      \\
        \hline
              28 &  Thermal conductivity & 1429      \\
         \hline
            29 &    Water flux & 1371      \\
         \hline
            30 &    Density & 1354      \\
         \hline
            31 &    Current density & 1337      \\
         \hline
         32 &    Limiting Oxygen Index  & 1146      \\
         \hline
         33 &    Transmittance & 1135      \\
         \hline
         34 &    Ion exchange capacity & 1034      \\
         \hline
         35 &    Porosity & 1019      \\
         \hline
         36 &    External quantum efficiency & 915      \\
         \hline
         37 &    Hole mobility & 846      \\
          \hline
         38 &    Capacity retention & 843      \\
         \hline
         39 &    Luminance & 842      \\
         \hline
         40 &   Compressive strength & 814      \\
         \hline
         41 &    Zeta potential & 784      \\
         \hline
         42 &    Sheet resistance & 768      \\
         \hline
         43 &    Lower critical solution temperature & 712  \\
         \hline
         44 &    CO\textsubscript{2} permeability & 685      \\
         \hline
         45 &    Resistivity & 668      \\
         \hline
         46 &    Coulombic efficiency & 632      \\
         \hline
         47 &    Crystallization temperature & 605      \\
         \hline
         48 &    Refractive index & 576      \\
         \hline
         49 & Separation factor & 543      \\
         \hline
         50 &    Impact strength & 512      \\
         \hline
         51 &    Highest occupied molecular orbital & 506 \\
         \hline
    \caption{The number of material property records extracted for the most common properties reported in the literature. This is not a complete list of extracted properties.}
    \label{tab:prop_records_elaborate}
    \end{longtable}

\section{S5. Training Machine learning models using literature extracted data}

\begin{figure}[!h]
    \raggedleft
    \subfigure[]{
		\begin{minipage}{0.46\textwidth}
			\includegraphics[width=1\textwidth]{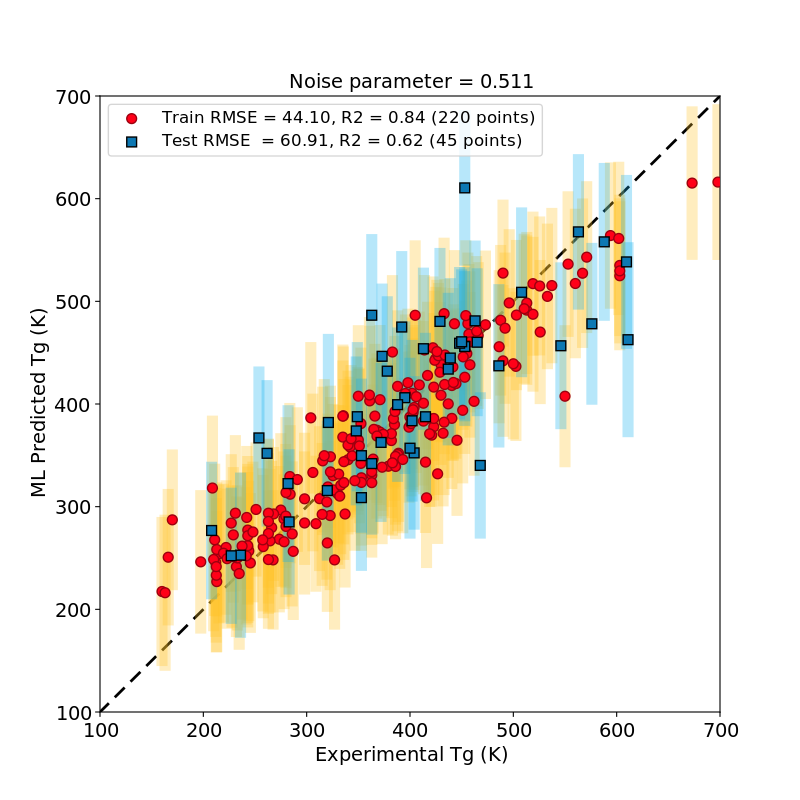}
		\end{minipage}
		\label{fig:extracted_ml}
	}
	\raggedright
	\subfigure[]{
		\begin{minipage}{0.46\textwidth}
			\includegraphics[width=1\textwidth]{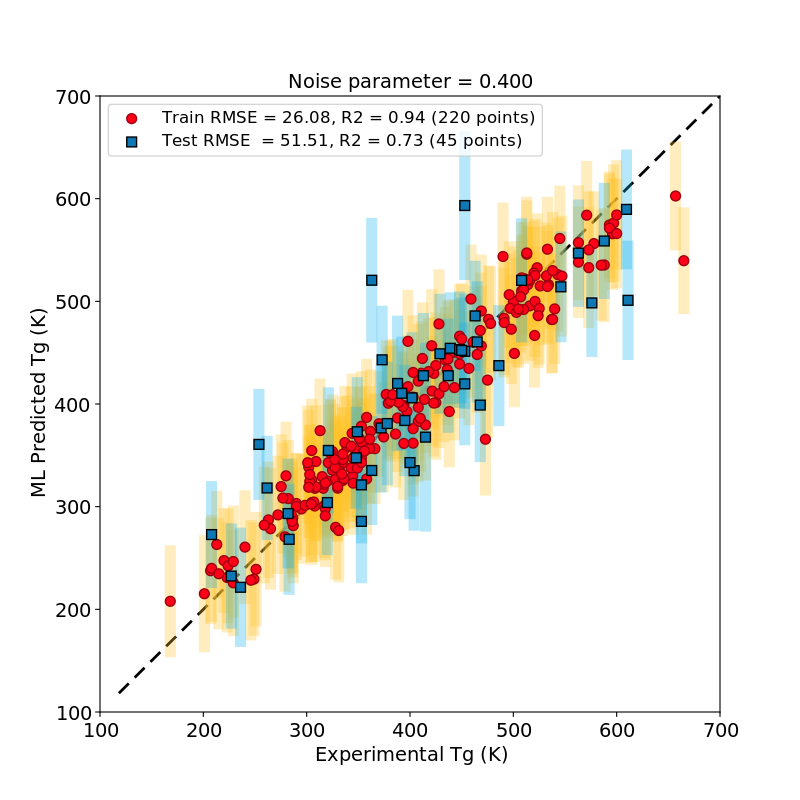}
		\end{minipage}
		\label{fig:curated_ml}
	}
	\caption{Glass Transition Temperature data a) Parity plot for machine learning model trained using literature extracted glass transition temperature data, b) Parity plot for machine learning model trained using curated glass transition temperature data with the same number of points as used in a). The test set in both cases is drawn from curated data. Observe that the test RMSE in both cases is similar}
% 	\vspace{-3mm}
	\label{fig:ml_predictions}
% 	\vspace{-3mm}
\end{figure}

In addition to the qualitative trends shown in the main paper that demonstrate the utility of our workflow, we also demonstrate how the data collected through our workflow can be used as inputs to machine learning models. We trained gaussian process regression models with a Matern kernel \cite{williams2006gaussian} to predict \Tg using the material property data extracted from our workflow. SMILES strings\cite{weininger1989smiles} were used to encode the structure of the polymer. The SMILES strings for a randomly selected subset of NLP extracted neat polymers \Tg records were added manually. The SMILES string was used as the input to fingerprint the polymer and the fingerprint vector along with the literature extracted \Tg value was the input to the machine learning model. The polymer SMILES string is converted to a feature vector using structural descriptors described elsewhere\cite{pg}. The feature vector consists of atomic triples, block level features, i.e., pre-defined fragments such as benzene rings and ketone groups, and chain level characteristics that are specific to a polymer such as length of longest side-chain, etc. This fingerprinting scheme leads to up to 600 non-zero fingerprint components.

Figure \ref{fig:ml_predictions}(a) shows the parity plot for a machine learning model to predict glass transition temperature trained using data extracted from literature. For comparison, we also trained a model using the same number of data points but with polymers and corresponding \Tg values randomly sampled from a curated \Tg data set reported in Ref. \citenum{pg} (Figure \ref{fig:ml_predictions}(b)). The test set is the same for both cases and is taken from the curated data set. There is an 80-20 split between the train and test set in both cases. The root mean squared error (RMSE)\cite{chai2014root} reported here on the test set is higher than the RMSE of the best \Tg models trained on larger data sets as reported in Ref. \citenum{pg} ($\sim$ 20 K). However, the test RMSE of the 2 plots in Figure \ref{fig:ml_predictions} is comparable, demonstrating that data extracted from literature can be used directly to train machine learning models of properties, without manual curation. Thus, despite being noisy, NLP extracted data in the context of organics and polymers can be used to train machine learning models which is important for NLP extracted data to be used at scale in materials informatics.

\end{document}